\documentclass[letterpaper]{article} % DO NOT CHANGE THIS
\usepackage[preprint]{aaai2027}  % DO NOT CHANGE THIS
\usepackage[hyphens]{url}  % DO NOT CHANGE THIS
\usepackage{graphicx} % DO NOT CHANGE THIS
\usepackage{natbib}  % DO NOT CHANGE THIS AND DO NOT ADD ANY OPTIONS TO IT
\usepackage{caption} % DO NOT CHANGE THIS AND DO NOT ADD ANY OPTIONS TO IT
\usepackage{algorithm}
\usepackage{algorithmic}

\usepackage{newfloat}
\usepackage{listings}
\DeclareCaptionStyle{ruled}{labelfont=normalfont,labelsep=colon,strut=off} % DO NOT CHANGE THIS
\floatstyle{ruled}
\newfloat{listing}{tb}{lst}{}
\floatname{listing}{Listing}

\usepackage{booktabs}

\usepackage[utf8]{inputenc} % allow utf-8 input
\usepackage[T1]{fontenc}    % use 8-bit T1 fonts
\usepackage{url}            % simple URL typesetting
\usepackage{booktabs}       % professional-quality tables
\usepackage{amsfonts}       % blackboard math symbols
\usepackage{nicefrac}       % compact symbols for 1/2, etc.
\usepackage{microtype}      % microtypography
\usepackage{xcolor}         % colors
\usepackage{amsmath}
\usepackage{tabularx}
\usepackage{adjustbox}
\usepackage{array}
\usepackage{graphicx}
\usepackage[american]{babel}
\usepackage{microtype}
\usepackage{subcaption}
\usepackage{enumitem}
\usepackage{makecell}
\usepackage{multirow}
\usepackage{tabularx}
\usepackage{array}
\usepackage{dblfloatfix}
\usepackage{svg}
\newcolumntype{C}{>{\centering\arraybackslash}X}

\usepackage[table]{xcolor}

\usepackage{float}
\usepackage{listings}
\usepackage{caption}
\usepackage{listings}
\DeclareCaptionStyle{ruled}{labelfont=normalfont,labelsep=colon,strut=off} % DO NOT CHANGE THIS
\floatstyle{ruled}
\newfloat{listing}{tb}{lst}{}
\floatname{listing}{Listing}
\title{Learning to Predict Middle-Layer Attention in MLLMs for Visual Token Pruning}
\author{
  Yuyao Sun\textsuperscript{\rm 1}\equalcontrib,
  Tao Deng\textsuperscript{\rm 1}\equalcontrib,
  Shuang Li\textsuperscript{\rm 1}\corresponding,
  Deqing Wang\textsuperscript{\rm 1}\corresponding,
  Hao Geng\textsuperscript{\rm 1},
  Minjun Yu\textsuperscript{\rm 2}
}
\affiliations{
    \textsuperscript{\rm 1}Beihang University\\
    \textsuperscript{\rm 2}Shanghai EABOT Technology Co., Ltd.
}

\usepackage{bibentry}
\begin{document}

\maketitle


\begin{abstract}

Multimodal large language models (MLLMs) achieve strong performance across diverse vision-language tasks, but their efficiency is limited by the cost of processing numerous visual tokens. Visual token pruning can reduce this cost, but requires accurate token importance estimates. Recent studies have demonstrated that text-to-vision attention from middle language model layers can effectively guide visual token pruning, typically using attention from a predefined middle layer to select the visual tokens to retain. Two problems therefore remain. First, our analysis shows that the layer whose attention is most responsive to the question varies substantially across samples, making a fixed layer suboptimal. Second, obtaining attention from the appropriate middle layer requires processing numerous visual tokens through several language model layers, by which point considerable computation has already been spent. To address both problems, we propose Middle-layer Attention Prediction (MAP), which uses Question Contrastive Teacher Selection to identify a sample-specific teacher layer by contrasting attention under the original and reference questions, and distills attention from the selected layer into a lightweight predictor that estimates visual token importance from multimodal input features. During inference, MAP combines the predicted importance scores with a diversity criterion to prune visual tokens before the first language model layer. Thus, MAP requires no attention maps for pruning and remains compatible with existing inference acceleration techniques. Across ten benchmarks on LLaVA-NeXT-7B, MAP retains $97.5\%$ of the unpruned model's performance with only $5.56\%$ of the visual tokens, yielding a $3.09\times$ end-to-end speedup.

\end{abstract}

\section{Introduction}
Multimodal large language models (MLLMs) have demonstrated strong capabilities across diverse vision-language tasks~\cite{flamingo,blip2,instructblip,llava,llava_1_5}. However, processing visual inputs remains computationally expensive, as each image is typically represented by hundreds or even thousands of visual tokens. To preserve fine-grained visual information, recent MLLMs often represent visual inputs with substantially more visual tokens~\cite{llavanext,llavaonevision,internvl_1_5,deepseek-vl2,internvl3,qwen3vl}, further increasing computational and memory costs. Consequently, visual token pruning has emerged as a promising approach to improving MLLM inference efficiency.

\begin{figure}[t]
    \centering

    \begin{subfigure}[t]{0.51\linewidth}
        \centering
        \includegraphics[width=\linewidth]{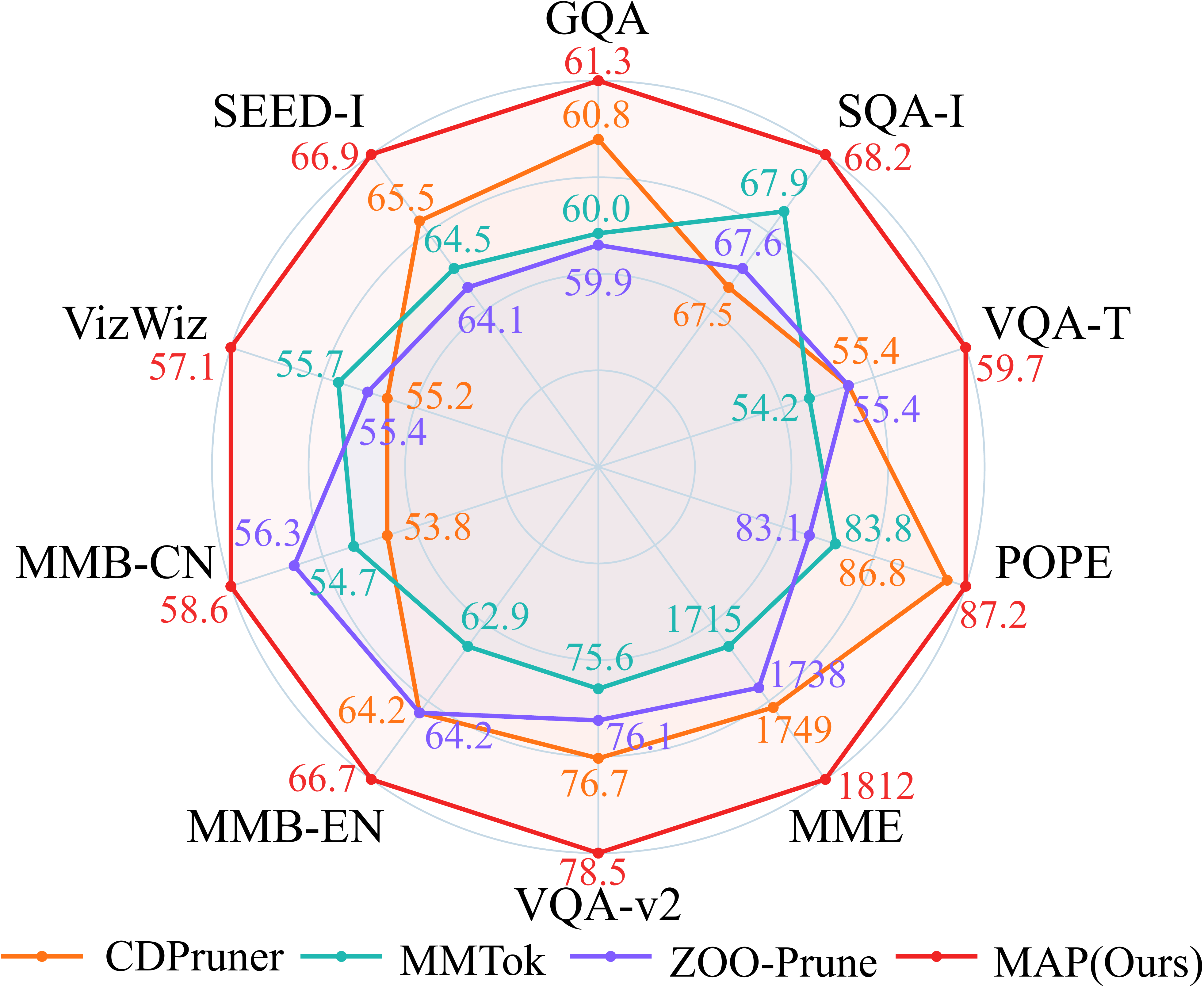}
        \captionsetup{skip=-0.0em}
        \caption{Benchmark performance}
        \label{fig:intro_performance}
    \end{subfigure}
    \hspace{-0.2em}
    \begin{subfigure}[t]{0.47\linewidth}
        \centering
        \includegraphics[width=\linewidth]{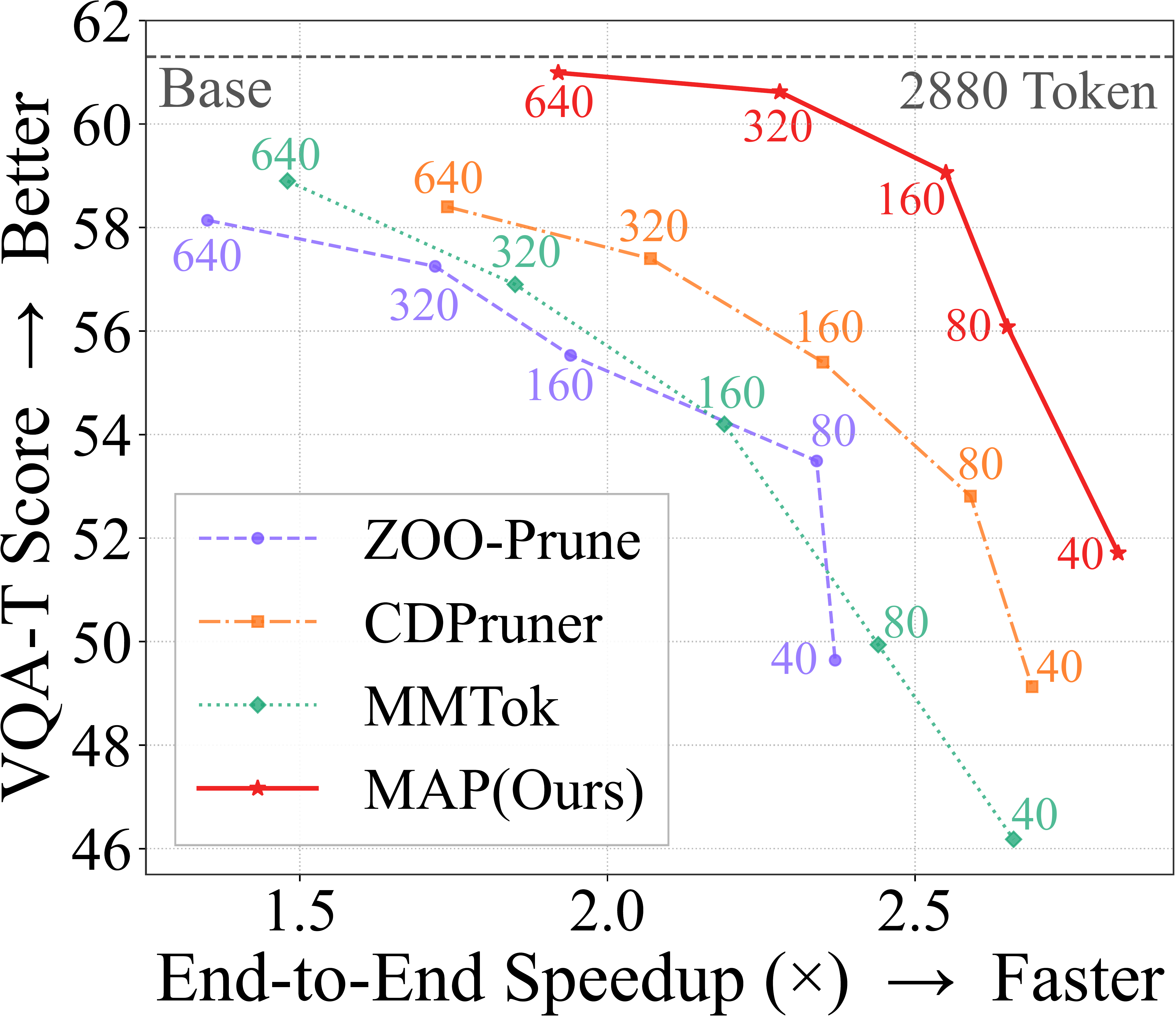}
        \captionsetup{skip=-0.0em}
        \caption{Accuracy--speed trade-off}
        \label{fig:intro_efficiency}
    \end{subfigure}
    \vspace{-0.2em}
    \caption{
        Performance and inference efficiency comparison on LLaVA-NeXT-7B.
        (a) Performance across ten vision-language benchmarks with 160
        retained visual tokens.
        (b) TextVQA accuracy versus end-to-end speedup at different retention
        levels; numbers denote retained visual tokens.
    }
    \label{fig:overview}
    \vspace{-1em}
\end{figure}

A prominent line of work uses text-to-vision attention within the language model to guide visual token pruning~\cite{fastv,sparsevlm,mustdrop}. However, recent studies have questioned whether attention scores reliably reflect token importance, showing that attention-guided pruning can even underperform random pruning at high compression ratios~\cite{vispruner,dart,visionzip}. LearnPruner~\cite{learnpruner} revisits this issue and shows that text-to-vision attention from middle language model layers can effectively guide visual token pruning. Specifically, it performs question-aware pruning in its second stage using attention extracted from a predefined middle layer. However, whether a fixed middle layer can provide effective pruning guidance across different inputs remains unclear.

To investigate this issue, we first analyze how text-to-vision attention localizes question-relevant visual evidence across language model layers. Using paired questions that refer to different objects in the same image, we find that attention from middle layers most clearly distinguishes regions relevant to the current question. More importantly, the layer providing the strongest localization varies substantially across samples. To assess whether these differences in localization are also reflected in pruning performance, we further evaluate layer-wise attention as a guidance signal for visual token pruning before LLM processing. We find that attention from middle layers generally provides the strongest pruning guidance, while the best layer differs across benchmarks. These findings suggest that, although middle-layer attention is broadly effective for pruning, no single layer is universally optimal.

However, exploiting middle-layer attention poses a fundamental efficiency dilemma. Although it provides effective pruning guidance, obtaining the signal requires numerous visual tokens to pass through several language model layers, by which point a substantial portion of the computation has already been incurred. Moreover, explicitly materializing attention maps during inference can limit compatibility with optimized attention kernels, further reducing efficiency.

To address the variation in the optimal layer across samples and the cost of obtaining middle-layer attention, we propose \textbf{Middle-layer Attention Prediction (MAP)}. During offline training, \textbf{Question Contrastive Teacher Selection (QCTS)} compares text-to-vision attention distributions for the input question and a generic reference question on the same image. Since layers insensitive to the question tend to produce similar attention for both questions, QCTS selects the layer with the largest difference as the teacher for each sample. 
A lightweight predictor is then trained to match the selected attention distribution through distribution matching, while the original MLLM remains frozen throughout training.

During inference, the predictor estimates the sample-specific text-to-vision attention distribution from multimodal representations available before language model processing. MAP uses the predicted attention scores to assess visual token importance, combines these scores with a diversity criterion, and prunes tokens before the first language model layer, allowing only the retained tokens to participate in subsequent computation. By avoiding reliance on LLM attention maps during inference, MAP remains compatible with optimized attention kernels such as FlashAttention~\cite{flashattention}.

Extensive experiments across multiple MLLM backbones, benchmarks, and visual token retention ratios validate the effectiveness of MAP. MAP consistently achieves a better accuracy--efficiency trade-off than existing visual token pruning methods. Across ten benchmarks on LLaVA-NeXT-7B, MAP retains $97.5\%$ performance using only $5.56\%$ of visual tokens. At this retention ratio, MAP achieves a $7.44\times$ speedup in the LLM prefill stage and a $3.09\times$ speedup in end-to-end inference. Our contributions are summarized as follows:   
\begin{itemize}
\item We show that middle-layer text-to-vision attention effectively guides pruning, while the layer most responsive to the question varies across samples.
\item To avoid the cost of obtaining this attention during inference, we propose \textbf{MAP}, which distills it into a lightweight predictor operating before language model processing.
\item We introduce QCTS, which contrasts attention distributions induced by the input and reference questions and selects the most question-responsive layer as a sample-specific teacher to supervise the predictor.
\item Extensive experiments across MLLMs and benchmarks show that MAP achieves a better accuracy--efficiency trade-off than existing visual token pruning methods.
\end{itemize}

\vspace{-1ex}

%------------------------------------------------------------------------
\section{Related Work}
\subsection{Multimodal Large Language Models}
Multimodal large language models (MLLMs) extend large language models to process visual inputs by connecting pretrained vision encoders with language models through cross-modal interfaces, including cross-attention layers~\cite{flamingo}, Q-Former~\cite{blip2,instructblip}, and lightweight projectors~\cite{llava,llava_1_5}.
To enhance visual perception and understanding, recent MLLMs encode visual inputs at a finer granularity to preserve detailed spatial and semantic information, producing long input sequences~\cite{llavanext,llavaonevision,internvl3,seedvl,qwen3vl}.
Processing a large number of visual tokens introduces substantial computational overhead, leading to the development of visual token pruning methods.

\subsection{Visual Token Pruning in MLLMs}

Visual token pruning improves MLLM efficiency by reducing the number of processed visual tokens while preserving performance. Many methods estimate token importance using attention scores~\cite{fastv,sparsevlm,fitprune}, although recent studies question their reliability under aggressive compression~\cite{vispruner,dart}. This has motivated alternative criteria such as instruction-conditioned diversity~\cite{cdpruner}, multimodal coverage~\cite{mmtok}, and token sensitivity~\cite{zoo}.
LearnPruner~\cite{learnpruner} shows that text-to-vision attention is useful in middle layers but less effective in early and late layers, and uses a predefined middle layer for question-guided token selection. Extending this analysis, we identify two limitations: the most question-responsive layer varies across inputs, and middle-layer pruning still requires many visual tokens to traverse the preceding layers. MAP addresses both by using QCTS to select a sample-specific teacher layer through question contrast and distilling its attention into a predictor before the first LLM layer for efficient early pruning during inference.

\vspace{-1ex}

\section{Motivation}
\label{sec:insights}

\begin{figure*}[!t]
    \vspace{-5em}
    \centering
    \captionsetup[subfigure]{
        font=small,
        justification=centering,
        singlelinecheck=true,
        skip=2pt
    }

    \newlength{\figpanelheight}
    \setlength{\figpanelheight}{0.34\textheight}

    % ---------------- Left: (a) ----------------
    \begin{subfigure}[b][\figpanelheight][s]{0.695\textwidth}
        \centering

        \vfill
        \includegraphics[
            width=\linewidth,
            trim=0 5 0 5,
            clip
        ]{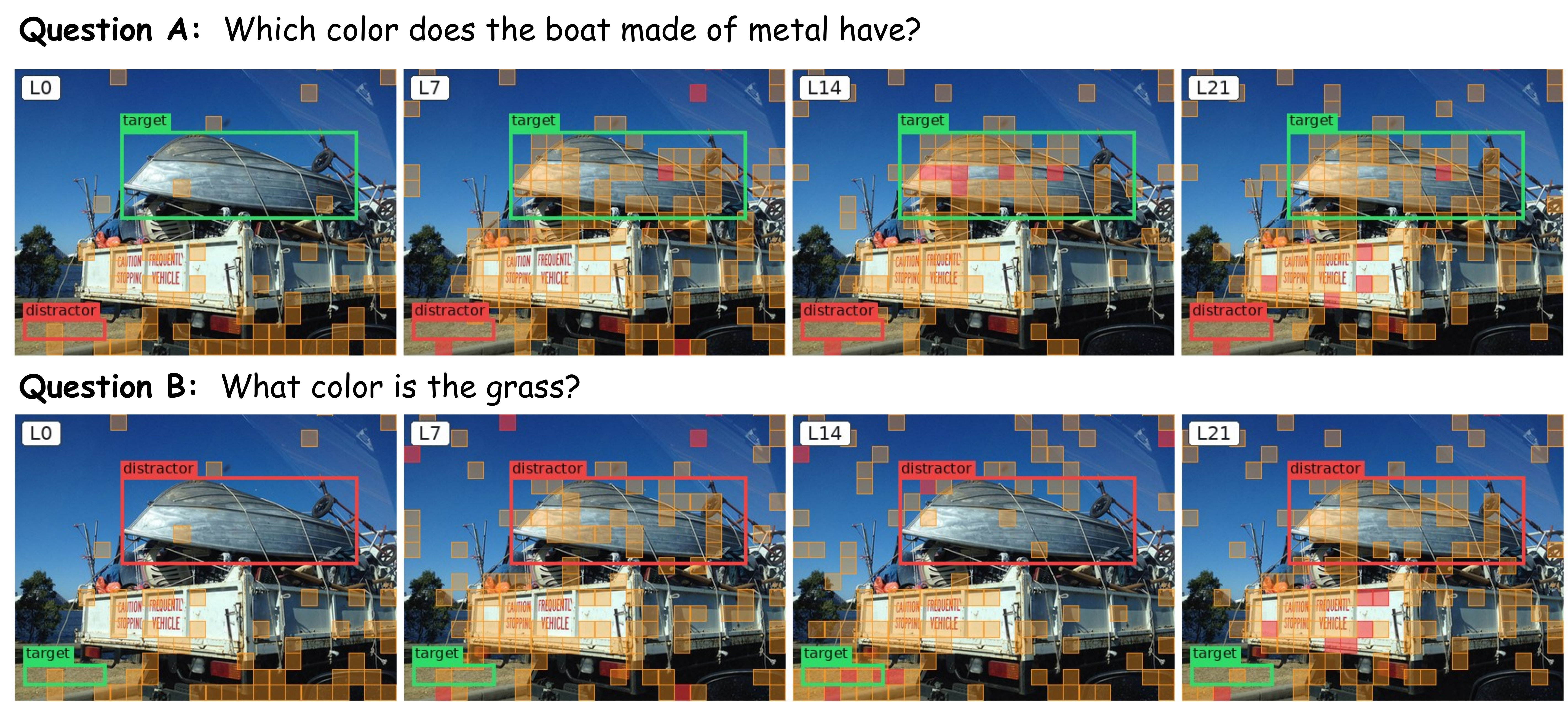}
        \vspace{-0.3em}

        \caption{Spatial distribution of visual tokens selected using text-to-vision attention across layers}
        \label{fig:case_attn_a}
    \end{subfigure}%
    \hfill%
    % ---------------- Right: (b) and (c) ----------------
    \begin{minipage}[b][\figpanelheight][b]{0.295\textwidth}
        \centering

        \begin{subfigure}[t]{\linewidth}
            \centering
            \includegraphics[width=\linewidth]
            {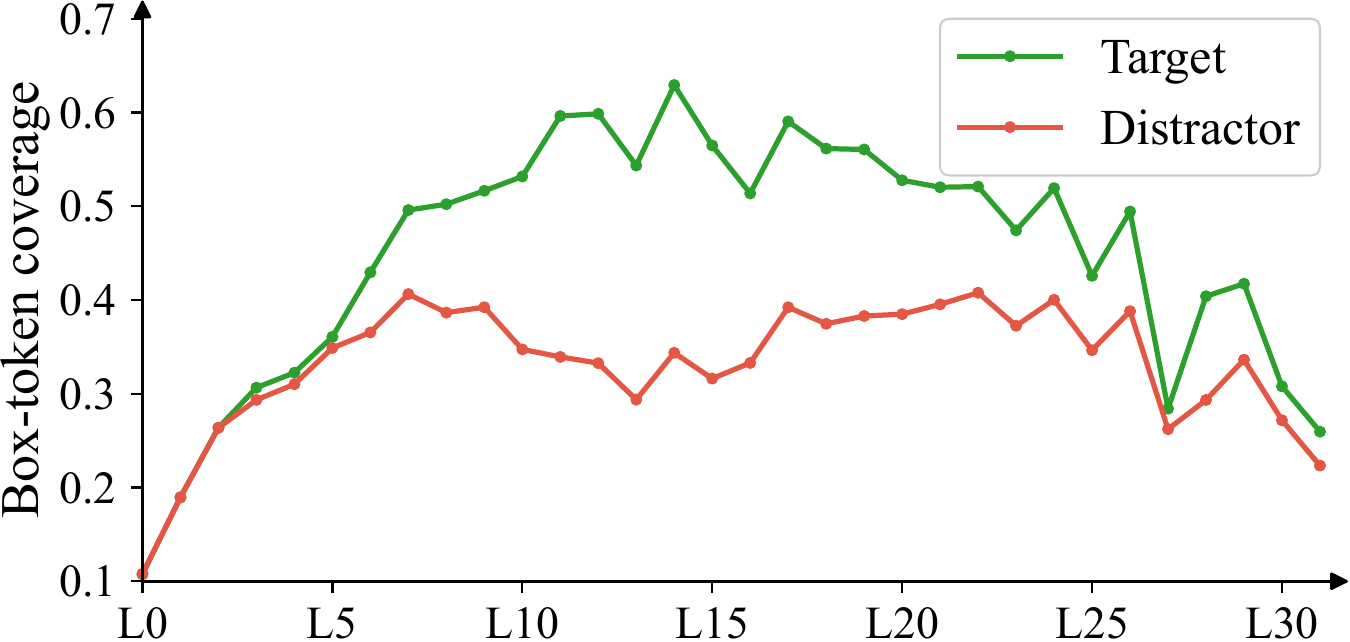}
            \caption{Token coverage of target and distractor}
            \label{fig:case_attn_b}
        \end{subfigure}

        \vspace{0.15em}

        \begin{subfigure}[b]{\linewidth}
            \centering
            \includegraphics[width=\linewidth]
            {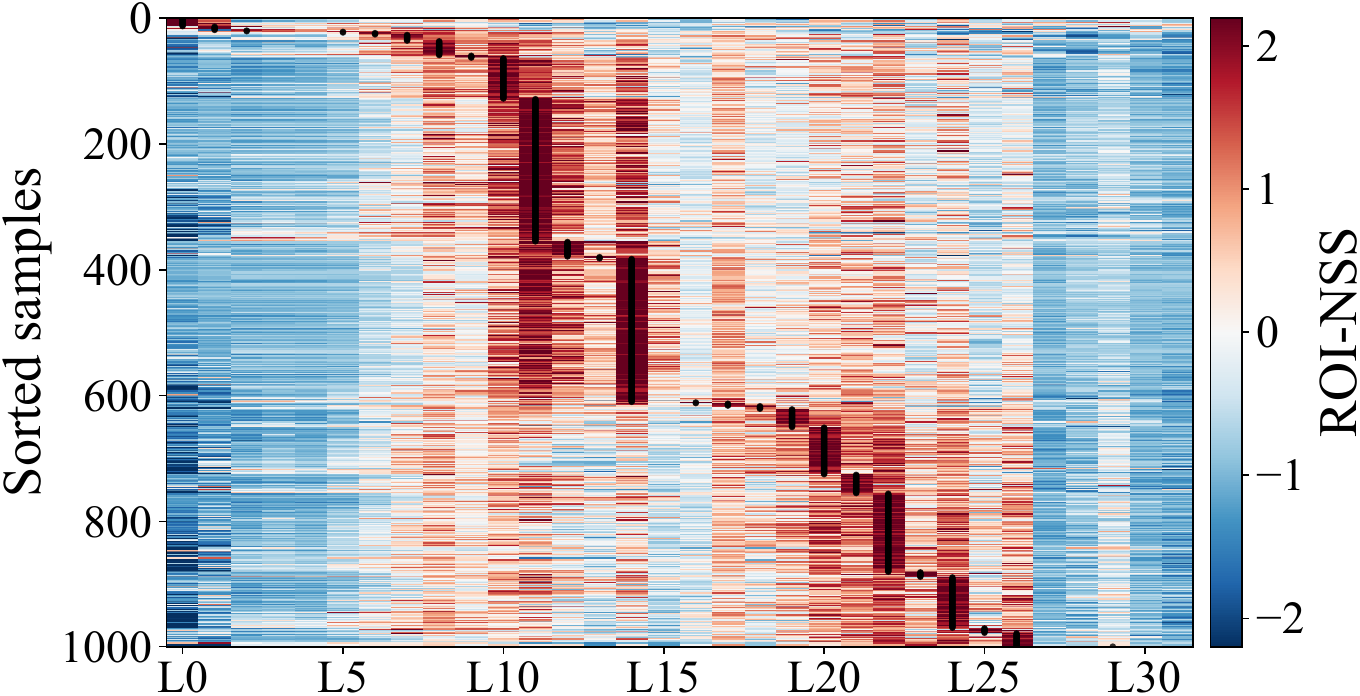}
            \caption{ROI-NSS across samples and layers}
            \label{fig:case_attn_c}
        \end{subfigure}
    \end{minipage}

    \vspace{-0.5em}
\caption{Analysis of how text-to-vision attention changes across language model layers. (a) The 128 visual tokens with the highest attention scores are highlighted. Target and distractor denote regions relevant and irrelevant to the question, respectively. (b) Average token coverage within the target and distractor boxes over 1000 questions. (c) ROI-NSS across questions and layers. ROI-NSS is the mean normalized attention score within the target box, with higher values indicating better localization.}
    \label{fig:case_attn}
\end{figure*}

\begin{figure*}[!t]
    \centering
    \includegraphics[
        width=0.98\textwidth
    ]{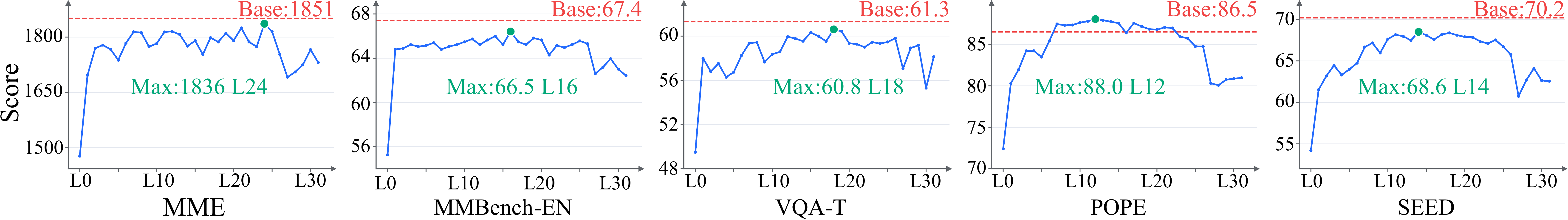}
    \caption{
Layer-wise evaluation of text-to-vision attention for visual token pruning on LLaVA-NeXT-7B. We report performance after pre-LLM pruning, retaining the top $11.1\%$ of visual tokens according to the attention scores from each layer.
    }
    \label{fig:layerwise_attn}
    \vspace{-1em}
\end{figure*}

\subsection{Question-Relevant Attention Across Layers}

Effective pruning requires preserving visual evidence relevant to the question. We therefore examine whether middle-layer attention more clearly localizes question-relevant targets and whether the layer providing the strongest localization is consistent across inputs. 
To isolate question effects, we construct 500 grounded question pairs from GQA~\cite{gqa} (1,000 questions), each sharing an image but referring to different objects. For each question, the referred object is the target and the other distractor. We then analyze the visual tokens top-ranked by attention across layers.

Fig.~\ref{fig:case_attn}(a) illustrates how attention localization changes across language model layers. At layer 0, token selection exhibits a strong positional bias, favoring tokens near the end of the sequence. At layer 7, the selections for the two questions remain similar despite their different targets. At layer 14, the selections become question-specific, focusing on the boat for Question A and the grass for Question B. This distinction weakens again in later layers.

To determine whether this pattern generalizes, Fig.~\ref{fig:case_attn}(b) reports the average coverage of the target and distractor objects by the top 128 visual tokens at each layer. In early layers, their coverage is similar, indicating limited distinction between the two objects. Toward the middle layers, target coverage increases while distractor coverage decreases, showing clearer localization of evidence relevant to the current question. In later layers, coverage declines for both objects. Together, Fig.~\ref{fig:case_attn}(a) and Fig.~\ref{fig:case_attn}(b) show that middle-layer attention most clearly distinguishes the current target from the distractor, both qualitatively and across the evaluated questions.

Fig.~\ref{fig:case_attn}(c) then examines whether the strongest localization occurs at the same layer across samples. ROI-NSS is computed by normalizing the attention scores across all visual tokens and averaging them within the target bounding box. For each question, the layer with the highest ROI-NSS is regarded as providing the strongest localization. Although higher ROI-NSS values are generally concentrated in the middle layers, the peak varies substantially across questions. Thus, no single predefined middle layer consistently provides the strongest question-relevant localization across inputs.

\subsection{Pruning Utility of Text-to-Vision Attention}
Having observed that the strongest question-relevant localization occurs at different layers across samples, we next show that the layer providing the best pruning guidance also differs across benchmarks. We evaluate this on five benchmarks with LLaVA-NeXT-7B by using attention from each language model layer to guide visual token pruning at layer 0. 
The unpruned model provides the baseline and layer-wise attention scores. Each layer's scores are then used in a separate forward pass that retains only the top $11.1\%$ of visual tokens. Fig.~\ref{fig:layerwise_attn} presents the results. Performance generally improves when attention is taken from early to middle layers and declines in later layers, while the best layer differs across benchmarks. With the best layer selected for each benchmark, the pruned model remains close to the unpruned baseline. These results show that middle-layer attention provides effective pruning guidance, but no single layer consistently performs best.

\begin{figure*}[t]
    \centering
    \includegraphics[width=\textwidth]{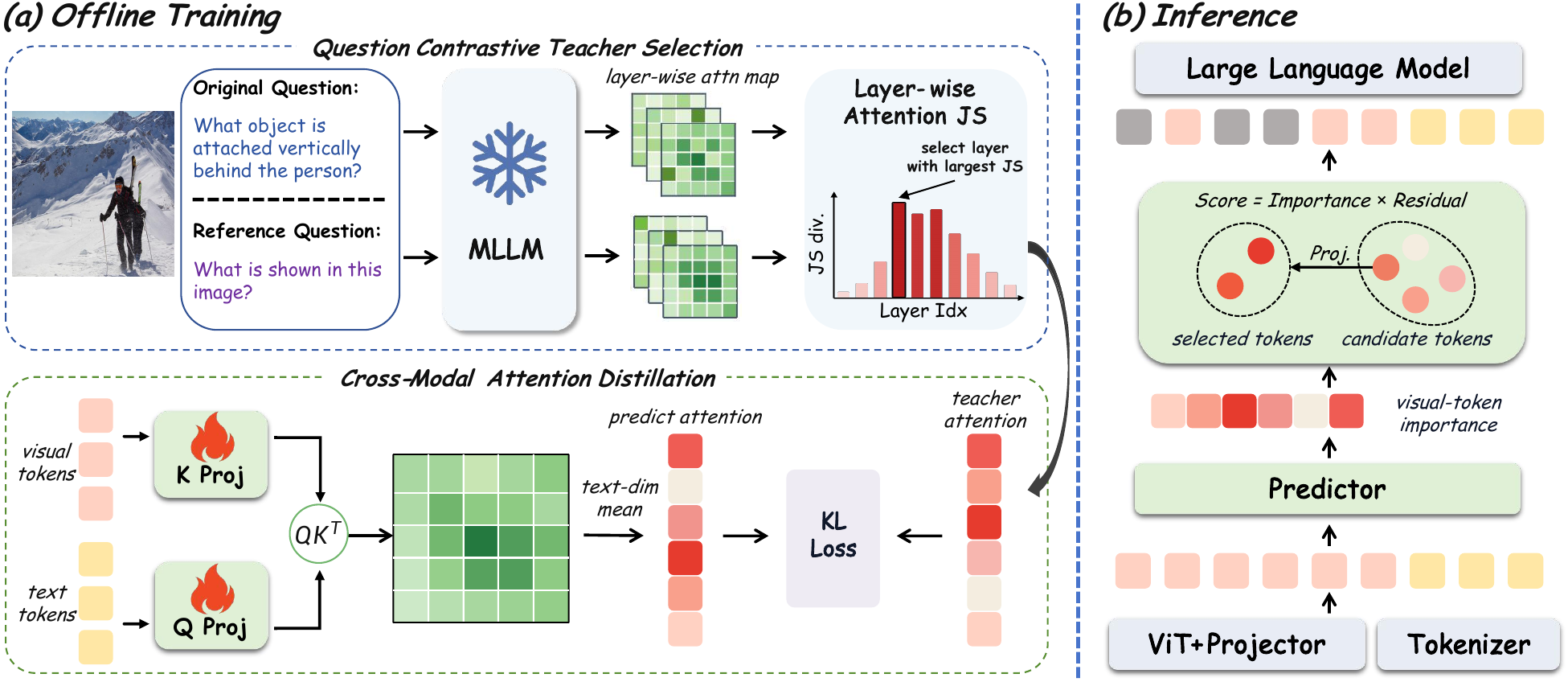}
    \caption{Overview of MAP.}
    \label{fig:framework}
    \vspace{-1em}
\end{figure*}

\vspace{-1ex}
%------------------------------------------------------------------------
\section{Method}\label{sec:method}

Our analysis shows that middle-layer text-to-vision attention provides effective guidance for visual token pruning. However, the layer most responsive to the question varies across samples, making a fixed-layer strategy suboptimal. Moreover, extracting this signal requires numerous visual tokens to pass through several language model layers, limiting the achievable efficiency gains. To overcome both limitations, we propose MAP, which distills text-to-vision attention from a sample-specific teacher layer into a lightweight predictor operating before language model processing.

%MAP consists of two phases: offline training and inference. During the offline phase, MAP first selects the middle layer whose visual attention is most sensitive to the target question as the teacher for each training sample. It then trains the predictor to reproduce the attention distributions produced by the selected teacher layers using the multimodal representations provided as input to the language model. During inference, MAP directly uses the predictor to estimate text-to-vision attention, treats the predicted attention as visual token importance, and combines this importance with visual feature diversity to select a compact token set that balances question relevance and information complementarity.

MAP consists of offline training and inference phases. During offline training, MAP selects the middle layer most responsive to the input question as the teacher for each sample and trains the predictor to reproduce its text-to-vision attention distribution from pre-LLM multimodal representations. During inference, the predictor directly estimates text-to-vision attention. The predicted attention scores serve as visual token importance, which MAP combines with visual feature diversity to retain a compact token set balancing question relevance and complementary information.

\begin{table*}[t]
\centering
\scriptsize
\renewcommand{\arraystretch}{0.94}
\setlength{\aboverulesep}{0.18ex}
\setlength{\belowrulesep}{0.18ex}
\setlength{\cmidrulesep}{0.10ex}
\begin{tabularx}{\textwidth}{l|*{10}{C}|C}
\toprule
Method
& GQA
& SQA$^{\scriptscriptstyle\mathrm{I}}$
& VQA$^{\scriptscriptstyle\mathrm{T}}$
& POPE
& MME
& VQA$^{\scriptscriptstyle\mathrm{v2}}$
& MMB$^{\scriptscriptstyle\mathrm{EN}}$
& MMB$^{\scriptscriptstyle\mathrm{CN}}$
& VizWiz
& SEED-I
& Rel.$\uparrow$ \\
\midrule

\rowcolor{gray!12}
\multicolumn{12}{c}{
    \textbf{\textit{All 576 Tokens}}
    \,
    \textbf{\textit{(100.0\%)}}
} \\
\midrule
Vanilla
& 61.9 & 69.5 & 58.2 & 85.9 & 1862
& 78.5 & 64.7 & 58.3 & 50.0 & 66.1 & 100.0\% \\

\midrule
\rowcolor{gray!12}
\multicolumn{12}{c}{
    \textbf{\textit{Retain 128 Tokens}}
    \,
    \textbf{\textit{($\downarrow$ 77.8\%)}}
} \\
\midrule
VisionZip (CVPR2025)
& 57.6 & 68.7 & 56.9 & 83.3 & 1761
& 75.6 & 62.1 & 57.0 & 51.6 & 61.2 & 96.7\% \\

DART (EMNLP2025)
& 57.9 & 69.1 & 56.3 & 80.4 & 1721
& 74.7 & 60.7 & 57.3 & 52.8 & 62.2 & 96.3\% \\

CDPruner (NeurIPS2025)
& 59.9 & 68.6 & 56.2 & 87.4 & 1745
& 76.6 & 63.1 & 55.0 & 52.8 & 63.2 & 97.8\% \\

MMTok (ICLR2026)
& 59.2 & 68.8 & 57.0 & 86.3 & 1779
& 76.3 & 62.3 & 55.1 & 52.9 & 63.1 & 97.8\% \\

ZOO-Prune (CVPR2026)
& 59.4 & 68.9 & 57.8 & 87.1 & 1751
& 76.5 & 61.8 & 55.6 & 52.1 & 62.8 & 97.7\% \\

%LearnPruner (ICLR2026)
%& 60.3 & 68.5 & 57.3 & 86.7 & 1820
%& 77.3 & 63.8 & 56.8 & -- & -- & 98.5 \\

\textbf{MAP (Ours)}
& 60.1 & 68.7 & 57.7 & 87.3 & 1800
& 76.9 & 63.1 & 57.0 & 52.5 & 64.4 & \textbf{98.9\%} \\

\midrule
\rowcolor{gray!12}
\multicolumn{12}{c}{
    \textbf{\textit{Retain 64 Tokens}}
    \,
    \textbf{\textit{($\downarrow$ 88.9\%)}}
} \\
\midrule
VisionZip (CVPR2025)
& 55.1 & 69.0 & 55.5 & 77.0 & 1690
& 72.4 & 60.1 & 55.4 & 52.9 & 57.3 & 93.7\% \\

DART (EMNLP2025)
& 54.7 & 69.3 & 54.7 & 73.8 & 1705
& 71.3 & 48.0 & 53.5 & 53.5 & 59.6 & 91.3\% \\

CDPruner (NeurIPS2025)
& 58.6 & 68.1 & 55.3 & 87.1 & 1717
& 75.4 & 61.1 & 53.2 & 53.4 & 62.1 & 96.4\% \\

MMTok (ICLR2026)
& 58.2 & 69.1 & 56.0 & 85.7 & 1715
& 75.2 & 61.2 & 54.5 & 53.9 & 61.4 & 96.6\% \\

LearnPruner (ICLR2026)
& 58.9 & 68.3 & 56.6 & 86.8 & 1750
& 76.0 & 62.6 & 55.7 & -- & -- & 96.9\% \\

ZOO-Prune (CVPR2026)
& 58.5 & 68.2 & 55.3 & 85.8 & 1657
& 75.0 & 60.2 & 54.7 & 54.0 & 60.9 & 95.9\% \\

\textbf{MAP (Ours)}
& 59.4 & 68.9 & 57.0 & 87.1 & 1744
& 76.1 & 63.2 & 57.0 & 52.8 & 62.9 & \textbf{98.1\%} \\

\midrule
\rowcolor{gray!12}
\multicolumn{12}{c}{
    \textbf{\textit{Retain 32 Tokens}}
    \,
    \textbf{\textit{($\downarrow$ 94.4\%)}}
} \\
\midrule
VisionZip (CVPR2025)
& 51.8 & 69.1 & 53.1 & 69.4 & 1536
& 67.1 & 57.0 & 50.3 & 52.4 & 53.1 & 88.3\% \\

DART (EMNLP2025)
& 52.9 & 69.3 & 52.2 & 69.1 & 1615
& 67.1 & 58.5 & 50.0 & 52.5 & 58.8 & 89.8\% \\

CDPruner (NeurIPS2025)
& 57.0 & 69.0 & 53.2 & 87.4 & 1657
& 73.6 & 59.6 & 49.6 & 53.1 & 60.8 & 94.3\% \\

MMTok (ICLR2026)
& 56.6 & 69.0 & 53.4 & 85.0 & 1635
& 73.4 & 59.4 & 49.8 & 54.6 & 59.7 & 93.9\% \\

LearnPruner (ICLR2026)
& 57.2 & 68.2 & 56.1 & 84.5 & 1672
& 74.0 & 60.8 & 55.5 & -- & -- & 94.8\%\\

ZOO-Prune (CVPR2026)
& 55.8 & 68.7 & 54.1 & 84.7 & 1643
& 72.2 & 58.9 & 50.7 & 53.7 & 58.2 & 93.4\% \\

\textbf{MAP (Ours)}
& 58.1 & 69.1 & 55.3 & 87.4 & 1726
& 74.6 & 62.2 & 54.6 & 52.9 & 61.5 & \textbf{96.6\%} \\
\bottomrule
\end{tabularx}
\caption{Performance comparison on LLaVA-1.5-7B with different numbers of retained visual tokens. Rel. denotes the average performance across benchmarks after normalizing each score by the corresponding vanilla score.}
\label{tab:llava15_results}
\vspace{-1em}
\end{table*}

\subsection{Question Contrastive Teacher Selection}
The utility of attention from different middle layers for visual token pruning varies across samples, making a fixed layer suboptimal as a universal teacher. To construct teacher supervision tailored to each sample, we propose Question Contrastive Teacher Selection (QCTS), which compares the text-to-vision attention produced for the input question with that produced for a generic reference question on the same image and then selects the layer with the largest discrepancy as the teacher. The selected teacher attention is expected to best capture visual regions relevant to the input question.

Given an image and question pair, the image is processed by the vision encoder and multimodal projector to produce $N$ visual token embeddings $\mathbf{V}=[\mathbf{v}_1,\ldots,\mathbf{v}_N]$, while the $M$ question tokens are mapped by the LLM embedding layer to text token embeddings $\mathbf{T}=[\mathbf{t}_1,\ldots,\mathbf{t}_M]$. Together, $\mathbf{V}$ and $\mathbf{T}$ constitute the multimodal input representations provided to the language model. Let $\mathcal{L}=\{0,\ldots,L-1\}$ denote the set of all $L$ language model layers. At each layer $l\in\mathcal{L}$, we extract text-to-vision attention and average it over all $H$ attention heads and $M$ question tokens for each visual token:
\begin{equation}
a_n^{l}
=
\frac{1}{HM}
\sum_{h=1}^{H}
\sum_{m=1}^{M}
A_{h,m,n}^{l}.
\end{equation}
Here, $A_{h,m,n}^{l}$ denotes the attention from question token $\mathbf{t}_m$ to visual token $\mathbf{v}_n$ at head $h$ of layer $l$. At each layer, we apply $\ell_1$ normalization to the averaged attention scores over visual tokens, yielding the target attention distributions $\{\mathbf{p}^{l}\}_{l\in\mathcal{L}}$.

We obtain the reference attention distributions by pairing the same image with the fixed reference question, ``What is shown in this image?'', while keeping the remaining input unchanged. Applying the attention aggregation and normalization procedure produces $\mathbf{p}_{\mathrm{ref}}^{l}$ at each layer, with attention averaged over the $M_{\mathrm{ref}}$ reference question tokens. As the reference question does not specify a particular object, attribute, relation, or region, it provides a generic visual attention baseline. The discrepancy between $\mathbf{p}^{l}$ and $\mathbf{p}_{\mathrm{ref}}^{l}$ therefore quantifies how the input question redirects visual attention relative to this baseline at layer $l$. Specifically, we quantify the discrepancy at each layer using the Jensen--Shannon divergence and select the layer with the largest value:
\begin{equation}
l^{\star}
=
\operatorname*{arg\,max}_{l\in\mathcal{L}}
D_{\mathrm{JS}}
\left(
\mathbf{p}^{l},
\mathbf{p}_{\mathrm{ref}}^{l}
\right).
\end{equation}
The attention distribution $\mathbf{p}^{l^{\star}}$ from the selected layer is then used as supervision to train the predictor.

%\subsection{Cross-Modal Attention Distillation}
%To estimate visual-token importance before language model processing, we train a lightweight predictor using the selected attention distribution $\mathbf{p}^{l^{\star}}$ from the teacher layer as supervision. The predictor takes the pre-LLM visual and question token embeddings $V$ and $T$ as input and outputs a visual-token importance distribution $\widehat{\mathbf{p}}=[\hat{p}_1,\ldots,\hat{p}_N]$.
%
%The predictor consists of two modality-specific MLPs, $f_T$ and $f_V$, followed by a lightweight scoring module containing only query and key projections. The two MLPs first map the question and visual tokens into a shared feature space. Let $\mathbf{W}_Q$ and $\mathbf{W}_K$ denote the query and key projection matrices, respectively. The scoring module projects the question features as queries and the visual features as keys. It then computes the importance of each visual token by averaging their dot-product similarities over all question tokens:
%\begin{equation}
%\hat{p}_n
%=
%\mathrm{softmax}_{n}
%\left(
%\operatorname{Avg}_{m}
%\left\langle
%\mathbf{W}_{Q}f_T(t_m),
%\mathbf{W}_{K}f_V(v_n)
%\right\rangle
%\right).
%\end{equation}
%Since the predictor only estimates token importance, it does not require value or output projections. During training, we minimize the KL divergence from $\mathbf{p}^{l^{\star}}$ to $\widehat{\mathbf{p}}$. This supervision distills the relative importance of all visual tokens from the selected teacher layer into the predictor.

\subsection{Cross-Modal Attention Distillation}

After QCTS selects the teacher distribution $\mathbf{p}^{l^\star}$ for each training sample, we train a lightweight predictor to reproduce it directly from the multimodal input representations of the LLM. The predictor first transforms the text and visual representations using two modality-specific MLPs, $f_T$ and $f_V$, and then projects them into queries and keys using $\mathbf{W}_Q$ and $\mathbf{W}_K$, respectively. Before computing the attention scores, we apply RoPE to the projected queries and keys to preserve positional information. The predicted attention distribution over visual tokens is computed as
%After QCTS selects the teacher distribution $\mathbf{p}^{l^\star}$ for each training sample, we train a lightweight predictor to reproduce it directly from the multimodal input representations of the LLM. The predictor consists of two modality-specific MLPs, $f_T$ and $f_V$, followed by a self-attention module with learnable query and key projections $\mathbf{W}_Q$ and $\mathbf{W}_K$. Before computing attention, we apply RoPE to the projected queries and keys to preserve positional information. The predicted attention distribution over visual tokens is computed as
\begin{equation}
\hat{p}_n
=
\frac{1}{M}
\sum_{m=1}^{M}
\operatorname{softmax}_{n}
\left(
\frac{
\left\langle
\mathbf{W}_Q f_T(\mathbf{t}_m),
\mathbf{W}_K f_V(\mathbf{v}_n)
\right\rangle
}{\sqrt{d}}
\right),
\end{equation}
where $d$ is the projection dimension. Since only attention weights are required, value and output projections are omitted.

During offline training, the predictor takes the multimodal input embeddings as input and is optimized by minimizing $D_{\mathrm{KL}}(\mathbf{p}^{l^\star}\|\widehat{\mathbf{p}})$. This objective distills the visual token importance from the selected teacher distribution into the predictor.

\subsection{Inference-Time Token Pruning}

During inference, MAP treats the predicted attention $\widehat{\mathbf{p}}$ as visual token importance. Given a target number $K$ of visual tokens to retain, MAP first keeps the top $K_0$ tokens with the highest predicted attention to form the initial retained set $\mathcal{R}_0$, while the remaining tokens form the candidate set $\mathcal{C}_0$. Starting from this initialization, MAP greedily adds one candidate token at each iteration until $K$ tokens are retained, jointly considering predicted importance and feature diversity.

To promote diversity, MAP maintains an orthonormal basis constructed from the selected visual embeddings. The initial basis $\mathbf{B}_0$ is obtained by orthonormalizing the embeddings in $\mathcal{R}_0$ and is updated whenever a new token is retained. At iteration $t$, the next token is selected by
\begin{equation}
i_t
=
\operatorname*{arg\,max}_{i \in \mathcal{C}_{t-1}}
\widehat{p}_i
\left\|
\mathbf{v}_i
-
\mathbf{B}_{t-1}
\mathbf{B}_{t-1}^{\top}
\mathbf{v}_i
\right\|_2.
\end{equation}
The importance term favors tokens related to the input question, while the orthogonal residual rewards features complementary to the current subspace. After $i_t$ is selected, it is moved from the candidate set to the retained set. Its embedding $\mathbf{v}_{i_t}$ is then orthogonalized against $\mathbf{B}_{t-1}$, and the normalized residual is appended to form $\mathbf{B}_t$. This procedure continues until the retained set contains $K$ tokens. Before language model processing, the selected tokens are restored to their original order while preserving their positional indices and are then combined with the question tokens.

\vspace{-1ex}

%------------------------------------------------------------------------
\section{Experiments}\label{sec:experiments}
\begin{table*}[t]
\centering
\scriptsize
\renewcommand{\arraystretch}{0.94}
\setlength{\aboverulesep}{0.18ex}
\setlength{\belowrulesep}{0.18ex}
\setlength{\cmidrulesep}{0.10ex}
\begin{tabularx}{\textwidth}{l|*{10}{C}|C}
\toprule
Method
& GQA
& SQA$^{\scriptscriptstyle\mathrm{I}}$
& VQA$^{\scriptscriptstyle\mathrm{T}}$
& POPE
& MME
& VQA$^{\scriptscriptstyle\mathrm{v2}}$
& MMB$^{\scriptscriptstyle\mathrm{EN}}$
& MMB$^{\scriptscriptstyle\mathrm{CN}}$
& VizWiz
& SEED-I
& Rel.$\uparrow$ \\
\midrule

\rowcolor{gray!12}
\multicolumn{12}{c}{
    \textbf{\textit{Upper Bound, All 2880 Tokens}}
    \,
    \textbf{\textit{(100.0\%)}}
} \\
\midrule
Vanilla
& 64.2 & 70.1 & 61.3 & 86.5 & 1851
& 81.8 & 67.4 & 60.6 & 57.6 & 70.2 & 100.0\% \\

\midrule
\rowcolor{gray!12}
\multicolumn{12}{c}{
    \textbf{\textit{Retain 640 Tokens}}
    \,
    \textbf{\textit{($\downarrow$ 77.8\%)}}
} \\
\midrule
VisionZip (CVPR2025)
& 61.2 & 68.1 & 59.9 & 86.0 & 1787
& 79.1 & 65.8 & 58.1 & 57.1 & 66.7 & 97.1\% \\

DART (EMNLP2025)
& 61.3 & 68.2 & 59.5 & 85.0 & 1721
& 78.3 & 64.9 & 57.1 & 57.0 & 67.9 & 96.3\% \\

CDPruner (NeurIPS2025)
& 62.6 & 67.9 & 58.4 & 87.3 & 1800
& 79.9 & 66.3 & 57.5 & 55.6 & 68.6 & 97.3\% \\

MMTok (ICLR2026)
& 62.2 & 68.4 & 58.9 & 86.7 & 1829
& 79.3 & 65.2 & 56.5 & 55.8 & 67.7 & 97.0\% \\

ZOO-Prune (CVPR2026)
& 62.2 & 67.7 & 58.0 & 86.7 & 1783
& 79.6 & 65.2 & 57.2 & 55.2 & 67.9 & 96.6\% \\

\textbf{MAP (Ours)}
& 62.8 & 68.9 & 61.0 & 88.2 & 1838
& 80.4 & 68.4 & 59.6 & 57.6 & 69.5 & \textbf{99.4\%} \\

\midrule
\rowcolor{gray!12}
\multicolumn{12}{c}{
    \textbf{\textit{Retain 320 Tokens}}
    \,
    \textbf{\textit{($\downarrow$ 88.9\%)}}
} \\
\midrule
VisionZip (CVPR2025)
& 58.9 & 67.5 & 58.8 & 82.5 & 1702
& 76.2 & 63.3 & 55.6 & 56.2 & 63.4 & 93.8\% \\

DART (EMNLP2025)
& 59.5 & 67.5 & 57.6 & 81.0 & 1705
& 75.7 & 64.2 & 55.7 & 56.8 & 64.8 & 93.9\% \\

CDPruner (NeurIPS2025)
& 61.6 & 67.8 & 57.4 & 87.2 & 1807
& 78.4 & 65.5 & 55.7 & 55.8 & 67.1 & 96.2\% \\

MMTok (ICLR2026)
& 60.9 & 67.3 & 56.9 & 85.7 & 1799
& 77.6 & 64.3 & 55.8 & 55.4 & 66.2 & 95.3\% \\

LearnPruner (ICLR2026)
& 62.2 & 68.6 & 58.4 & -- & 1845
& 78.3 & 66.8 & -- & -- & --
& 97.4\% \\

ZOO-Prune (CVPR2026)
& 61.0 & 67.2 & 57.3 & 85.5 & 1787
& 78.1 & 64.9 & 56.4 & 55.1 & 66.5 & 95.5\% \\

\textbf{MAP (Ours)}
& 61.8 & 68.6 & 60.6 & 87.8 & 1842
& 79.6 & 67.1 & 59.0 & 57.5 & 68.4 & \textbf{98.5\%} \\

\midrule
\rowcolor{gray!12}
\multicolumn{12}{c}{
    \textbf{\textit{Retain 160 Tokens}}
    \,
    \textbf{\textit{($\downarrow$ 94.4\%)}}
} \\
\midrule
VisionZip (CVPR2025)
& 55.2 & 67.9 & 55.0 & 75.8 & 1630
& 71.4 & 58.6 & 50.4 & 55.5 & 58.3 & 88.5\% \\

DART (EMNLP2025)
& 56.8 & 67.8 & 54.9 & 75.3 & 1615
& 72.5 & 62.0 & 53.6 & 56.7 & 60.6 & 90.3\% \\

CDPruner (NeurIPS2025)
& 60.8 & 67.5 & 55.4 & 86.8 & 1749
& 76.7 & 64.2 & 53.8 & 55.2 & 65.5 & 94.3\% \\

MMTok (ICLR2026)
& 60.0 & 67.9 & 54.2 & 83.8 & 1715
& 75.6 & 62.9 & 54.7 & 55.7 & 64.5 & 93.3\% \\

LearnPruner (ICLR2026)
& 58.7 & 67.6 & 55.0 & -- & 1784
& 76.2 & 65.3 & -- & -- & --
& 94.0\% \\

ZOO-Prune (CVPR2026)
& 59.9 & 67.6 & 55.4 & 83.1 & 1738
& 76.1 & 64.2 & 56.3 & 55.4 & 64.1 & 93.9\% \\

\textbf{MAP (Ours)}
& 61.3 & 68.2 & 59.7 & 87.2 & 1812
& 78.5 & 66.7 & 58.6 & 57.1 & 66.9 & \textbf{97.5\%} \\
\bottomrule
\end{tabularx}
\caption{Performance comparison on LLaVA-NeXT-7B with different numbers of retained visual tokens.}
\label{tab:llavanext_results}
\vspace{-1em}
\end{table*}

\noindent\textbf{Datasets and Models.}
To empirically validate the effectiveness and generalizability of MAP, we conduct comprehensive experiments on three widely used MLLM backbones: LLaVA-1.5-7B~\cite{llava_1_5}, LLaVA-NeXT-7B~\cite{llavanext}, and Qwen2.5-VL-7B-Instruct~\cite{qwen2.5vl}. For LLaVA-1.5-7B and LLaVA-NeXT-7B, we evaluate MAP on GQA~\cite{gqa}, SQA~\cite{sqa}, TextVQA~\cite{textvqa}, POPE~\cite{pope}, MME~\cite{mme}, VQAv2~\cite{vqav2}, MMBench~\cite{mmbench}, VizWiz~\cite{vizwiz}, and SEED-Bench~\cite{seedbench}. For Qwen2.5-VL-7B-Instruct, we report results on AI2D~\cite{ai2d}, MME, MMBench, TextVQA, and GQA.

\noindent\textbf{Compared Methods.}
%We compare MAP with representative visual token pruning methods. VisionZip~\cite{visionzip} selects dominant visual tokens using attention scores from the ViT encoder. DART~\cite{dart} prunes redundant tokens based on their similarity to a small set of pivot tokens. CDPruner~\cite{cdpruner} employs an instruction-conditioned DPP objective to balance relevance and diversity. MMTok~\cite{mmtok} formulates token selection as a multimodal maximum coverage problem, jointly covering text and visual tokens. ZOO-Prune~\cite{zoo} estimates token importance through sensitivity analysis.
%We compare MAP with representative token pruning methods. FastV~\cite{fastv} prunes visual tokens using early-layer attention scores. VisionZip~\cite{visionzip} selects dominant tokens using ViT attention. DART~\cite{dart} removes redundant tokens by similarity to pivot tokens. CDPruner~\cite{cdpruner} balances relevance and diversity with an instruction-conditioned DPP objective. MMTok~\cite{mmtok} formulates token selection as a multimodal maximum coverage problem. ZOO-Prune~\cite{zoo} estimates token importance through sensitivity analysis.
We compare MAP with representative token pruning methods. FastV~\cite{fastv} prunes visual tokens using early layer attention. VisionZip~\cite{visionzip} selects tokens using ViT attention. DART~\cite{dart} removes redundant tokens by similarity to pivot tokens. CDPruner~\cite{cdpruner} balances relevance and diversity with an instruction-conditioned DPP. MMTok~\cite{mmtok} formulates token selection as a multimodal maximum coverage problem. ZOO-Prune~\cite{zoo} estimates token importance through sensitivity analysis. LearnPruner~\cite{learnpruner} removes visual redundancy using learned importance and diversity, then prunes query-irrelevant tokens based on middle-layer text-to-vision attention.

\iffalse
\begin{table}[h]
\centering
\renewcommand{\arraystretch}{0.94}
\caption{Performance comparison on Qwen2.5-VL-7B under different token budgets.}
\label{tab:qwen_results}
\begin{tabular}{@{}l|ccccc|c@{}}
\toprule
Method & AI2D & MME & VQA$^{\scriptscriptstyle\mathrm{T}}$ & GQA & POPE & Rel.$\uparrow$ \\
\midrule
\rowcolor{gray!12}
\multicolumn{7}{c}{\textbf{All 1280 Tokens}} \\
\midrule
Vanilla
& 83.2 & 2325 & 83.0 & 61.1 & 88.0 & 100 \\
\midrule

\rowcolor{gray!12}
\multicolumn{7}{c}{\textbf{(I) 512 Tokens}} \\
\midrule
FastV
& 81.2 & 2377 & 81.4 & 59.9 & 88.1 & 94.3 \\
%CDPruner
%& 82.1 & 2342.0 & 79.5 & 77.0 & 67.5 & 92.0 \\
VisionZip
& 82.7 & 2337 & 80.8 & 60.1 & 88.2 & 95.6 \\
MAP
& 82.8 & 2350 & 82.2 & 60.8 & 88.3 & 97.8 \\
\midrule

\rowcolor{gray!12}
\multicolumn{7}{c}{\textbf{(II) 256 Tokens}} \\
\midrule
FastV
& 76.9 & 2242 & 76.2 & 56.7 & 86.1 & 80.3 \\
%CDPruner
%& 81.0 & 2292.8 & 75.1 & 61.0 & 51.4 & 82.9 \\
VisionZip
& 81.3 & 2339 & 74.2 & 60.3 & 87.2 & 88.6 \\
MAP
& 81.0 & 2348 & 79.7 & 59.9 & 88.3 & 92.7 \\
\midrule

\rowcolor{gray!12}
\multicolumn{7}{c}{\textbf{(III) 128 Tokens}} \\
\midrule
FastV
& 68.2 & 1745 & 58.3 & 52.1 & 79.0 & 54.0 \\
%CDPruner
%& 78.8 & 2184.3 & 68.0 & 44.1 & 35.8 & 72.3 \\
VisionZip
& 77.5 & 2168 & 60.4 & 58.0 & 85.5 & 72.2 \\
MAP
& 79.4 & 2298 & 75.1 & 58.6 & 87.5 & 84.9 \\
\bottomrule
\end{tabular}
\end{table}
\fi

\begin{table}[t]
\centering
\scriptsize
\renewcommand{\arraystretch}{0.9}
\setlength{\aboverulesep}{0.18ex}
\setlength{\belowrulesep}{0.18ex}
\setlength{\cmidrulesep}{0.10ex}
\begin{tabularx}{\columnwidth}{l|*{6}{C}|C}
\toprule
Method
& AI2D
& MME
& MMB$^{\scriptscriptstyle\mathrm{EN}}$
& MMB$^{\scriptscriptstyle\mathrm{CN}}$
& VQA$^{\scriptscriptstyle\mathrm{T}}$
& GQA
& Rel.$\uparrow$ \\
\midrule

\rowcolor{gray!12}
\multicolumn{8}{c}{
    \textbf{\textit{All 1280 Tokens}}
    \,
    \textbf{\textit{(100.0\%)}}
} \\
\midrule
Vanilla
& 83.2 & 2325 & 83.7 & 80.5 & 83.0 & 61.1 & 100.0\% \\
\midrule

\rowcolor{gray!12}
\multicolumn{8}{c}{
    \textbf{\textit{Retain 512 Tokens}}
    \,
    \textbf{\textit{($\downarrow$ 60\%)}}
} \\
\midrule
FastV
& 78.8 & 2317 & 81.5 & 79.1 & 81.4 & 58.4 & 97.3\% \\
VisionZip
& 82.7 & 2337 & 81.8 & 79.7 & 80.8 & 60.1 & 98.7\% \\
\textbf{MAP}
& 82.8 & 2350 & 82.9 & 79.8 & 82.2 & 60.8 & \textbf{99.6\%} \\
\midrule

\rowcolor{gray!12}
\multicolumn{8}{c}{
    \textbf{\textit{Retain 256 Tokens}}
    \,
    \textbf{\textit{($\downarrow$ 80\%)}}
} \\
\midrule
FastV
& 76.2 & 2238 & 78.8 & 74.4 & 76.2 & 53.6 & 92.3\% \\
VisionZip
& 79.3 & 2209 & 79.7 & 76.6 & 74.2 & 57.3 & 94.0\% \\
\textbf{MAP}
& 81.0 & 2348 & 82.2 & 78.0 & 79.7 & 59.9 & \textbf{97.9\%} \\
\midrule

\rowcolor{gray!12}
\multicolumn{8}{c}{
    \textbf{\textit{Retain 128 Tokens}}
    \,
    \textbf{\textit{($\downarrow$ 90\%)}}
} \\
\midrule
FastV
& 68.2 & 1745 & 69.5 & 65.3 & 58.3 & 52.1 & 79.4\% \\
VisionZip
& 72.5 & 2168 & 76.0 & 74.5 & 60.4 & 54.0 & 87.5\% \\
\textbf{MAP}
& 79.4 & 2298 & 80.3 & 76.3 & 75.1 & 58.6 & \textbf{95.2\%} \\
\bottomrule
\end{tabularx}
\caption{Performance comparison on Qwen2.5-VL-7B.}
\label{tab:qwen_results}
\vspace{-1em}
\end{table}

\subsection{Main Results.}
%Tables~\ref{tab:llava15_results} and~\ref{tab:llavanext_results} report the overall comparisons on LLaVA-1.5-7B and LLaVA-NeXT-7B, respectively. MAP achieves the best average performance across all evaluated settings on both backbones. When retaining only \textbf{22.2\%}, 11.1\%, and \textbf{5.6\%} of the original visual tokens, MAP outperforms the strongest baselines by \textbf{0.8}, \textbf{1.3}, and \textbf{2.1} points on LLaVA-1.5-7B, and by \textbf{2.0}, \textbf{2.6}, and \textbf{2.6} points on LLaVA-NeXT-7B. The advantage becomes more pronounced as the retention ratio decreases, indicating that MAP preserves question-relevant visual evidence more effectively under aggressive pruning. Its consistent gains on LLaVA-NeXT-7B further demonstrate its effectiveness on longer visual sequences, where token selection is more challenging.

Tables~\ref{tab:llava15_results} and~\ref{tab:llavanext_results} report comparisons on LLaVA-1.5-7B and LLaVA-NeXT-7B, respectively. MAP achieves the best average performance across all settings on both backbones. When retaining only \textbf{22.2\%}, \textbf{11.1\%}, and \textbf{5.6\%} of the original visual tokens, MAP outperforms the strongest baselines by \textbf{1.1}, \textbf{1.2}, and \textbf{1.8} points on LLaVA-1.5-7B, and by \textbf{2.1}, \textbf{1.1}, and \textbf{3.2} points on LLaVA-NeXT-7B. The advantage becomes pronounced as the retention ratio decreases, indicating that MAP preserves question-relevant visual evidence more effectively under aggressive pruning. Its consistent gains on LLaVA-NeXT-7B demonstrate its effectiveness on longer visual sequences, where token selection is more challenging.

We further evaluate MAP on Qwen2.5-VL-7B-Instruct in Table~\ref{tab:qwen_results}. MAP consistently outperforms existing methods across all token retention ratios, improving the average relative score over the strongest baseline by \textbf{0.9} points with 512 tokens, \textbf{3.9} points with 256 tokens, and a substantially larger \textbf{7.7} points with 128 tokens. Notably, when restricted to only 128 tokens, existing methods degrade sharply, whereas MAP maintains an average relative score of \textbf{95.2\%}. These results demonstrate that MAP better preserves question-relevant visual evidence at high compression ratios.

\subsection{Ablation of Teacher Selection for Direct Pruning}

\begin{table}[t]
\centering
\scriptsize
\renewcommand{\arraystretch}{0.9}
\setlength{\tabcolsep}{1.5pt}
\setlength{\aboverulesep}{0.18ex}
\setlength{\belowrulesep}{0.18ex}
\setlength{\cmidrulesep}{0.10ex}
\begin{tabular*}{\columnwidth}{
    @{\extracolsep{\fill}}lcccccc@{}
}
\toprule
Method
& Cov.$\uparrow$
& ROI-NSS$\uparrow$
& Layer Dist.$\downarrow$
& Hit@1$\uparrow$
& Hit@3$\uparrow$
& Hit@5$\uparrow$ \\
\midrule

ROI-Max
& 66.2\%
& 2.728
& 0.0
& 100.0\%
& 100.0\%
& 100.0\% \\
\midrule

Random
& 56.5\%
& 1.210
& 6.5
& 5.8\%
& 16.0\%
& 26.3\% \\

Layer 14
& 61.3\%
& 2.056
& 4.6
& 22.8\%
& 42.5\%
& 52.7\% \\

\textbf{QCTS}
& \textbf{64.1\%}
& \textbf{2.421}
& \textbf{3.3}
& \textbf{47.8\%}
& \textbf{69.8\%}
& \textbf{79.8\%} \\

\bottomrule
\end{tabular*}
\caption{Comparison of layer-selection strategies.}
\label{tab:qcts_roi_best}
\vspace{-1em}
\end{table}

QCTS is designed to identify a teacher layer whose attention best captures the visual regions relevant to each question. To assess this ability, we use the same 1,000 image--question samples from the preceding analysis. ROI-Max selects the layer with the highest ROI-NSS for each question. Random uniformly selects one layer from Layers 8--24 for each question, while Layer 14 is the best-performing fixed baseline. For each method, we retain the top 128 visual tokens according to the attention scores from its identified layer. Macro Coverage measures the average target-region token coverage, ROI-NSS measures the concentration of attention within the target region, Layer Dist.\ and Hit@$K$ measure how closely the identified layer matches ROI-Max. As shown in Table~\ref{tab:qcts_roi_best}, QCTS consistently outperforms both baselines, achieving 64.1\% Macro Coverage, an ROI-NSS of 2.421, and a Layer Dist.\ of 3.3. Its Hit@1/3/5 rates reach 47.8\%, 69.8\%, and 79.8\%, respectively. The coverage and ROI-NSS demonstrate that QCTS preserves question-relevant visual information, while the low Layer Dist.\ and high Hit@$K$ rates show that it reliably identifies layers close to ROI-Max.

\begin{figure}[t]
    \centering
    \begin{subfigure}[t]{0.48\linewidth}
        \centering
        \includegraphics[width=\linewidth]{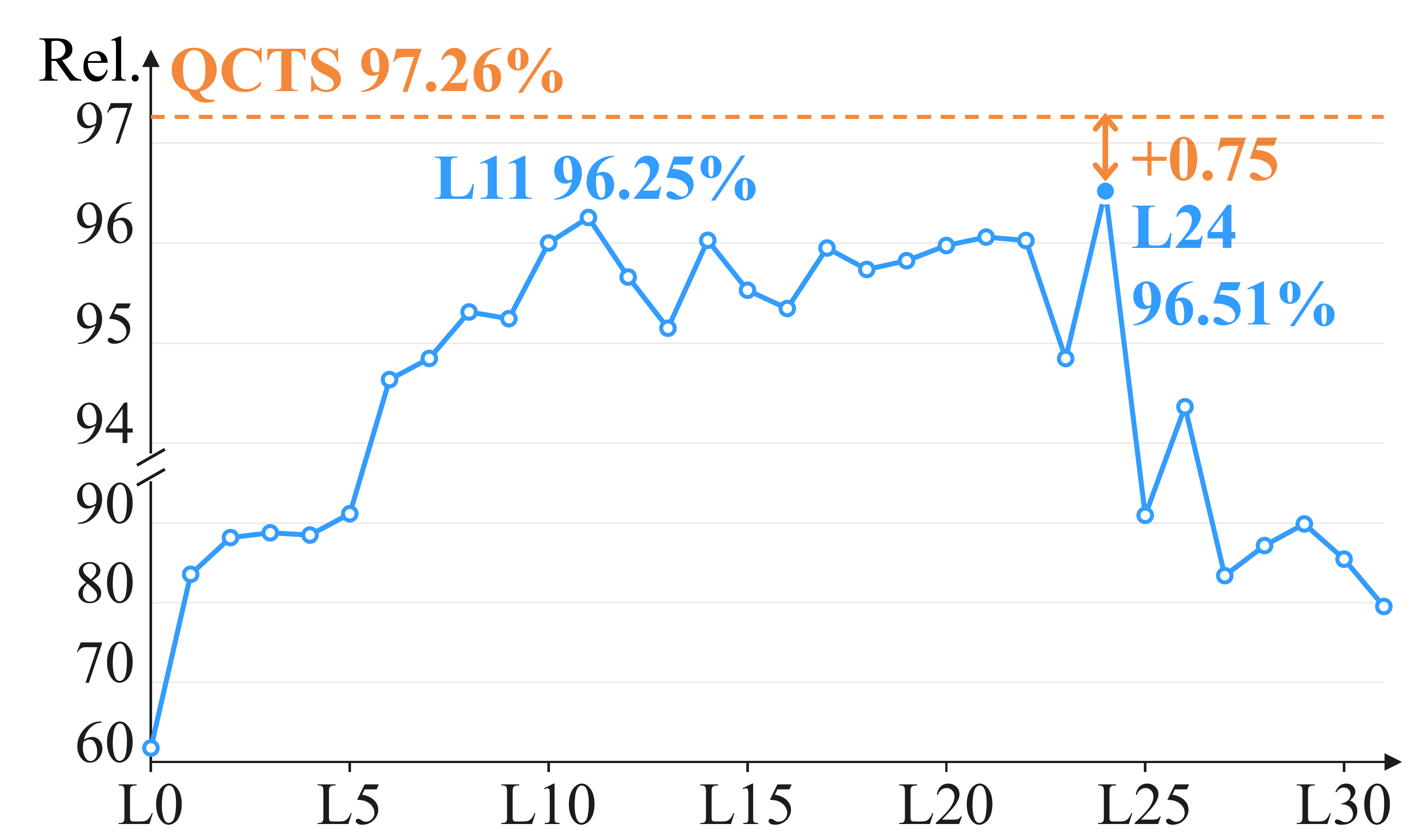}
        \caption{LLaVA-1.5-7B}
        \label{fig:example_a}
    \end{subfigure}
    \hfill
    \begin{subfigure}[t]{0.48\linewidth}
        \centering
        \includegraphics[width=\linewidth]{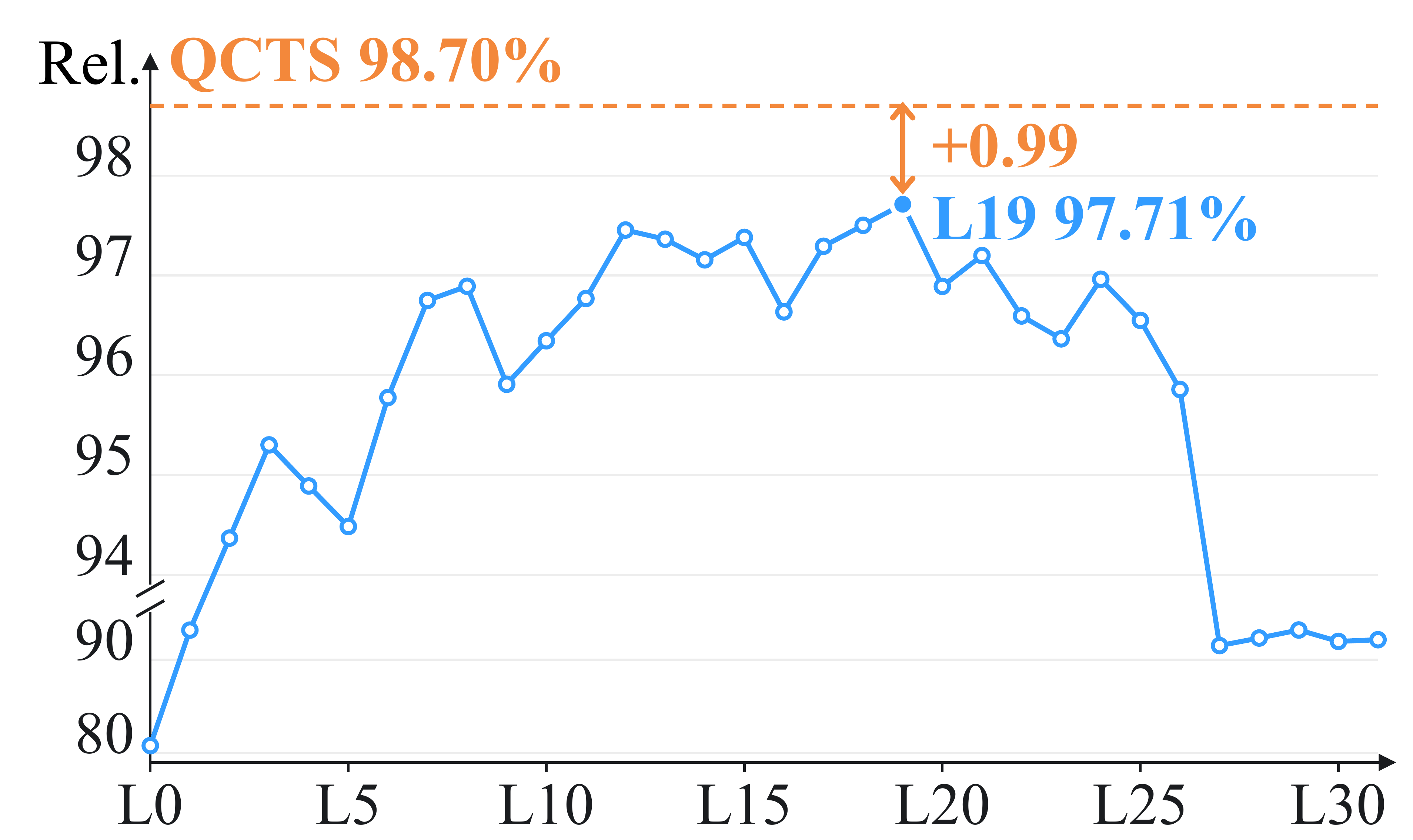}
        \caption{LLaVA-NeXT-7B}
        \label{fig:example_b}
    \end{subfigure}
    \vspace{-0.5em}
    \caption{Pruning performance of different layers and QCTS.}
    \label{fig:qualitative_comparison}
\end{figure}

We further evaluate whether the layers selected by QCTS provide effective signal for visual token pruning. Figure~\ref{fig:qualitative_comparison} compares it with attention from each fixed layer when only the top 11.1\% of visual tokens ranked by text-to-vision attention are retained. The y-axis represents the average relative performance across the six benchmarks in Table~\ref{tab:ablation_teacher}. QCTS achieves average relative performance of 97.26\% and 98.70\% on LLaVA-1.5-7B and LLaVA-NeXT-7B, exceeding the strongest fixed-layer results by 0.75 and 0.99 percentage points, respectively. These results show that attention from the layers identified by QCTS is more effective for visual token pruning than attention from any fixed layer.

%We further evaluate whether the layers selected by QCTS provide effective signal for visual token pruning. Figure~\ref{fig:qualitative_comparison} compares QCTS-selected attention with attention from each fixed layer for visual token pruning. The top 11.1\% of visual tokens ranked by text-to-vision attention are retained before entering the LLM, and the y-axis reports the average relative performance across the six benchmarks in Table~\ref{tab:ablation_teacher}. QCTS achieves 97.26\% and 98.70\% on LLaVA-1.5-7B and LLaVA-NeXT-7B, surpassing the strongest fixed-layer results by 0.75 and 0.99 percentage points, respectively, and demonstrating its effectiveness as a pruning signal.

\subsection{Ablation of Teacher Selection for Predictor Training}

\iffalse
\begin{table}[t]
\centering
\scriptsize
\renewcommand{\arraystretch}{1.02}
\setlength{\aboverulesep}{0.18ex}
\setlength{\belowrulesep}{0.18ex}
\setlength{\cmidrulesep}{0.10ex}
\caption{Ablation study on teacher attention selection. Results are reported on LLaVA-1.5-7B with 64 visual tokens.}
\vspace{-0.5em}
\resizebox{\columnwidth}{!}{%
\begin{tabular}{lcccccc}
\toprule
Teacher Source
& GQA
& TextVQA
& POPE
& MME
& MMB$^{\mathrm{EN/CN}}$
& Rel. \\
\midrule
Base
& 61.9 & 58.2 & 85.9 & 1862 & 64.7/58.3 & 100.0\% \\

\midrule
\rowcolor{gray!12}
\multicolumn{7}{c}{\textbf{(I) Single-layer teacher selection}} \\
\midrule
Fixed Layer 11
& 58.6 & 55.4 & 87.3(86.9) & 1702 & 60.9/56.4 & 95.6\% \\
Fixed Layer 14
& 59.0 & 55.9 & 87.1(86.5) & 1744 & 61.9/56.2 & 96.4\% \\
Fixed Layer 17
& 59.0 & 55.3 & 86.9(86.2) & 1698 & 60.8/56.4 & 95.6\% \\
QCTS-1
& 58.7 & 56.5 & 86.6 & 1757 & 62.4/56.4 & 96.7\% \\
\midrule
\rowcolor{gray!12}
\multicolumn{7}{c}{\textbf{(II) Multi-layer teacher fusion}} \\
\midrule
QCTS-2
& 58.9 & 57.0 & 86.8 & 1720 & 61.8/56.0 & 96.3\% \\
QCTS-4
& 58.9 & 56.5 & 86.6 & 1728 & 62.1/56.7 & 96.5\% \\
QCTS-8
& 59.1 & 56.8 & 86.6 & 1748 & 61.7/56.7 & 96.7\% \\
QCTS-1 + Layer 14
& 59.0 & 56.5 & 87.4 & 1757 & 62.2/56.5 & 96.9\% \\
\bottomrule
\end{tabular}%
}
\vspace{-1em}
\end{table}
\fi

\begin{table}[t]
\centering
\scriptsize
\renewcommand{\arraystretch}{0.94}
\setlength{\aboverulesep}{0.18ex}
\setlength{\belowrulesep}{0.18ex}
\setlength{\cmidrulesep}{0.10ex}
\resizebox{\columnwidth}{!}{%
\begin{tabular}{@{}l|cccccc@{}}
\toprule
Teacher Attn
& GQA
& TextVQA
& POPE
& MME
& MMB$^{\mathrm{EN/CN}}$
& Rel. \\
\midrule
Vanilla
& 61.9 & 58.2 & 85.9 & 1862 & 64.7/58.3 & 100.0\% \\

\midrule
\rowcolor{gray!12}
\multicolumn{7}{c}{\textbf{Single-layer teacher selection}} \\
\midrule
Fixed Layer 11
& 58.7 & 55.9 & 87.1 & 1757 & 62.5/56.3 & 96.6\% \\
Fixed Layer 14
& 59.0 & 56.9 & 86.8 & 1698 & 62.7/56.2 & 96.4\% \\
Fixed Layer 24
& 58.7 & 56.6 & 86.3 & 1745 & 63.1/55.4 & 96.5\% \\
QCTS
& 59.4 & 57.0 & 87.1 & 1744 & 63.2/57.0 & 97.4\% \\
\midrule
\rowcolor{gray!12}
\multicolumn{7}{c}{\textbf{Multi-layer teacher fusion}} \\
\midrule
QCTS-2
& 58.8 & 56.8 & 87.1 & 1735 & 62.6/56.3 & 96.7\% \\
QCTS-4
& 58.9 & 56.7 & 86.9 & 1737 & 63.0/56.4 & 96.9\% \\
QCTS-8
& 59.1 & 56.5 & 87.3 & 1739 & 63.2/56.0 & 96.9\% \\
QCTS-16
& 58.5 & 56.3 & 87.2 & 1726 & 62.5/55.9 & 96.3\% \\
\bottomrule
\end{tabular}%
}
\caption{Performance of predictors trained with different teacher-selection strategies on LLaVA-1.5-7B at 64 tokens.}
\label{tab:ablation_teacher}
\vspace{-1em}
\end{table}

We investigate the effect of teacher layer selection by training predictors with different strategies and comparing their pruning performance. A fixed-layer teacher uses attention from the same predefined layer as supervision for all training samples. We report Layers 11, 14, and 24, as they are the three best performing fixed layer teachers. As shown in Table~\ref{tab:ablation_teacher}, QCTS achieves the best overall pruning performance, with a relative score of 97.4\%, surpassing the strongest fixed-layer teacher by 0.8 points. This improvement suggests that QCTS can adaptively select a more suitable teacher layer for each sample, thereby providing more effective supervision for predictor training. We further evaluate QCTS-$n$, which averages the attention from the top $n$ layers selected by QCTS for each sample as teacher supervision. However, multi-layer fusion provides no additional gains and instead leads to a slight decrease in pruning performance. These results show that QCTS outperforms both fixed-layer selection and multi-layer teacher fusion. We therefore use QCTS to select the teacher layer for training the predictor in all experiments.

\begin{table}[t]
\centering
\scriptsize
\renewcommand{\arraystretch}{0.94}
\setlength{\aboverulesep}{0.18ex}
\setlength{\belowrulesep}{0.18ex}
\setlength{\cmidrulesep}{0.10ex}
\resizebox{\columnwidth}{!}{%
\begin{tabular}{@{}lccccccc@{}}
\toprule
Selection
& Tokens
& GQA
& TextVQA
& POPE
& MME
& MMB$^{\mathrm{EN/CN}}$
& Rel.$\uparrow$ \\
\midrule
Top-$K$ & 128
& 60.0 & 57.3 & 87.1 & 1789 & 63.3/56.8 & 98.0\% \\
+Diversity & 128
& 60.1 & 57.7 & 87.3 & 1800 & 63.1/57.0 & 98.3\% \\
\midrule
Top-$K$ & 64
& 59.0 & 56.4 & 87.1 & 1721 & 63.7/56.4 & 96.9\% \\
+Diversity & 64
& 59.4 & 57.0 & 87.1 & 1744 & 63.2/57.0 & 97.4\% \\
\midrule
Top-$K$ & 32
& 57.1 & 54.3 & 87.4 & 1642 & 62.8/53.9 & 94.2\% \\
+Diversity & 32
& 58.1 & 55.3 & 87.4 & 1726 & 62.2/54.6 & 95.5\% \\
\bottomrule
\end{tabular}%
}
\caption{Effect of diversity-aware selection with different numbers
of retained visual tokens on LLaVA-1.5-7B.}
\label{tab:selection_ablation}
\end{table}

%\subsection{Effect of Diversity-Aware Token Selection}
%Table~\ref{tab:selection_ablation} studies the contribution of diversity-aware selection under different numbers of retained visual tokens. The Top-K baseline directly retains the top-$K$ tokens according to the predicted attention scores, while the Diversity variant applies greedy re-ranking. At 128 tokens, Top-$K$ and Top-$K$ + Diversity achieve relative scores of 98.0\% and 98.3\%, respectively. This marginal improvement indicates that the predictor alone already captures a complete set of question-relevant visual evidence. As fewer tokens are retained, the diversity-aware variant improves the relative score from 96.9\% to 97.4\% at 64 tokens and from 94.2\% to 95.5\% at 32 tokens. These larger gains suggest that importance-based ranking may retain redundant visual information under stronger pruning, thereby limiting evidence coverage. By reducing such redundancy, it preserves more complementary evidence and becomes beneficial as the number of retained tokens decreases.

\subsection{Effect of Diversity-Aware Token Selection}
Table~\ref{tab:selection_ablation} studies the contribution of diversity-aware selection under different token budgets. The Top-$K$ baseline directly retains the $K$ tokens with the highest predicted attention scores, whereas the diversity-aware variant additionally considers feature complementarity. At 128 tokens, diversity-aware selection provides only a marginal improvement, from 98.0\% to 98.3\%. This result indicates that the predicted importance scores accurately identify the question-relevant tokens and already cover the key visual evidence under a sufficient token budget, leaving limited room for diversity to help. As the budget decreases, however, the improvement grows from 96.9\% to 97.4\% at 64 tokens and from 94.2\% to 95.5\% at 32 tokens. Under aggressive pruning, redundancy among highly ranked tokens becomes more costly. Diversity-aware selection mitigates this issue by preserving complementary visual features, leading to larger gains at smaller token budgets.

\subsection{Efficiency Analysis}

\begin{table}[t]
\centering
\scriptsize
\renewcommand{\arraystretch}{0.94}
\setlength{\aboverulesep}{0.18ex}
\setlength{\belowrulesep}{0.18ex}
\begin{tabular*}{\columnwidth}{@{\extracolsep{\fill}}lcccc@{}}
\toprule
Method & Tokens & Prefill (ms) & End-to-End (ms) & KV Cache (MB) \\
\midrule

\rowcolor{gray!12}
\multicolumn{5}{c}{\textbf{LLaVA-1.5-7B}} \\
\midrule
Vanilla
& 576
& 58.9
& 101.5
& 336 \\
%MAP
%& 128
%& 28.4 {\scriptsize(2.10$\times$)}
%& 101.6 {\scriptsize(1.35$\times$)}
%& 120 {\scriptsize($-65.1\%$)} \\
MAP
& 32
& 21.5 {\scriptsize(2.73$\times$)}
& 67.2 {\scriptsize(1.51$\times$)}
& 64 {\scriptsize(-80.95\%)} \\

\midrule
\rowcolor{gray!12}
\multicolumn{5}{c}{\textbf{LLaVA-NeXT-7B}} \\
\midrule
Vanilla
& 2880
& 224.5
& 277.9
& 1488 \\
%MAP
%& 640
%& 63.7 {\scriptsize(3.55$\times$)}
%& 168.9 {\scriptsize(1.92$\times$)}
%& 371 {\scriptsize($-74.7\%$)} \\
MAP
& 160
& 30.2 {\scriptsize(7.44$\times$)}
& 89.8 {\scriptsize(3.09$\times$)}
& 128 {\scriptsize(-91.39\%)} \\
\bottomrule
\end{tabular*}
\caption{Average inference efficiency across ten benchmarks.}
\label{tab:efficiency}
\vspace{-1em}
\end{table}

Table~\ref{tab:efficiency} reports average efficiency across ten benchmarks. MAP achieves up to 2.73$\times$ prefill speedup and 80.95\% KV cache reduction on LLaVA-1.5-7B, and 7.44$\times$ prefill and 3.09$\times$ end-to-end speedups with 91.39\% KV cache reduction on LLaVA-NeXT-7B. Detailed results for all ten benchmarks and a runtime breakdown are provided in the appendix.

\vspace{-1ex}
%------------------------------------------------------------------------
\section{Conclusion}
%In this work, we introduced MAP, which distills sample-specific middle layer attention into a lightweight predictor for input-side visual token pruning. During inference, MAP combines predicted saliency with diversity to retain a compact token set before LLM processing. Experiments across multiple backbones and benchmarks consistently demonstrate a favorable performance–efficiency trade-off.

%In this work, we show that attention from middle layers provides a reliable pruning signal, while the most informative layer varies across samples. MAP uses QCTS to supervise a lightweight predictor that estimates this attention before LLM processing, then combines the prediction with diversity for token pruning. Across three MLLM backbones, MAP achieves a favorable balance between accuracy and efficiency.

In this work, we show that the middle layer most responsive to the question varies across samples and its attention is costly to obtain. MAP addresses both issues by using QCTS to select a sample-specific teacher and distilling its attention into a lightweight predictor.  At inference, MAP combines predicted importance with feature diversity to prune visual tokens before the first LLM layer, without requiring attention maps. Experiments on three MLLM backbones demonstrate the favorable accuracy--efficiency trade-off of MAP.

%In this work, we show that the most question-responsive middle layer varies across samples and that obtaining its attention is costly. MAP addresses both issues by using QCTS to select a sample-specific teacher and distilling its attention into a lightweight predictor. At inference, MAP combines predicted importance with feature diversity to prune visual tokens before the first LLM layer, without requiring LLM attention maps. Experiments on three MLLM backbones demonstrate the favorable accuracy--efficiency trade-off of MAP.

\newpage
\bibliography{aaai2027}

@article{qwen2.5vl,
  title={Qwen2.5-VL Technical Report},
  author={Bai, Shuai and Chen, Keqin and Liu, Xuejing and Wang, Jialin and Ge, Wenbin and Song, Sibo and Dang, Kai and Wang, Peng and Wang, Shijie and Tang, Jun and others},
  journal={arXiv preprint arXiv:2502.13923},
  year={2025}
}

@article{internvl3,
  title={Internvl3: Exploring advanced training and test-time recipes for open-source multimodal models},
  author={Zhu, Jinguo and Wang, Weiyun and Chen, Zhe and Liu, Zhaoyang and Ye, Shenglong and Gu, Lixin and Tian, Hao and Duan, Yuchen and Su, Weijie and Shao, Jie and others},
  journal={arXiv preprint arXiv:2504.10479},
  year={2025}
}

@inproceedings{blip2,
  title={Blip-2: Bootstrapping language-image pre-training with frozen image encoders and large language models},
  author={Li, Junnan and Li, Dongxu and Savarese, Silvio and Hoi, Steven},
  booktitle={ICML},
  year={2023},
}

@inproceedings{instructblip,
  title={Instructblip: Towards general-purpose vision-language models with instruction tuning},
  author={Dai, Wenliang and Li, Junnan and Li, Dongxu and Tiong, Anthony and Zhao, Junqi and Wang, Weisheng and Li, Boyang and Fung, Pascale N and Hoi, Steven},
  booktitle={NeurIPS},
  year={2023}
}

@inproceedings{flamingo,
  title={Flamingo: a visual language model for few-shot learning},
  author={Alayrac, Jean-Baptiste and Donahue, Jeff and Luc, Pauline and Miech, Antoine and Barr, Iain and Hasson, Yana and Lenc, Karel and Mensch, Arthur and Millican, Katherine and Reynolds, Malcolm and others},
  booktitle={NeurIPS},
  year={2022}
}

@inproceedings{llava,
  title={Visual instruction tuning},
  author={Liu, Haotian and Li, Chunyuan and Wu, Qingyang and Lee, Yong Jae},
  booktitle={NeurIPS},
  year={2023}
}

@inproceedings{llava_1_5,
  title={Improved baselines with visual instruction tuning},
  author={Liu, Haotian and Li, Chunyuan and Li, Yuheng and Lee, Yong Jae},
  booktitle={CVPR},
  year={2024}
}

@misc{llavanext,
  title={Llava-next: Improved reasoning, ocr, and world knowledge},
  author={Liu, Haotian and Li, Chunyuan and Li, Yuheng and Li, Bo and Zhang, Yuanhan and Shen, Sheng and Lee, Yong Jae},
  year={2024}
}

@article{internvl_1_5,
  title={How far are we to gpt-4v? closing the gap to commercial multimodal models with open-source suites},
  author={Chen, Zhe and Wang, Weiyun and Tian, Hao and Ye, Shenglong and Gao, Zhangwei and Cui, Erfei and Tong, Wenwen and Hu, Kongzhi and Luo, Jiapeng and Ma, Zheng and others},
  journal={arXiv preprint arXiv:2404.16821},
  year={2024}
}

@article{sparsevlm,
  title={Sparsevlm: Visual token sparsification for efficient vision-language model inference},
  author={Zhang, Yuan and Fan, Chun-Kai and Ma, Junpeng and Zheng, Wenzhao and Huang, Tao and Cheng, Kuan and Gudovskiy, Denis and Okuno, Tomoyuki and Nakata, Yohei and Keutzer, Kurt and others},
  journal={arXiv preprint arXiv:2410.04417},
  year={2024}
}

@article{mustdrop,
  title={Multi-Stage Vision Token Dropping: Towards Efficient Multimodal Large Language Model},
  author={Liu, Ting and Shi, Liangtao and Hong, Richang and Hu, Yue and Yin, Quanjun and Zhang, Linfeng},
  journal={arXiv preprint arXiv:2411.10803},
  year={2024}
}

@inproceedings{textvqa,
  title={Towards vqa models that can read},
  author={Singh, Amanpreet and Natarajan, Vivek and Shah, Meet and Jiang, Yu and Chen, Xinlei and Batra, Dhruv and Parikh, Devi and Rohrbach, Marcus},
  booktitle={CVPR},
  year={2019}
}

@inproceedings{gqa,
  title={Gqa: A new dataset for real-world visual reasoning and compositional question answering},
  author={Hudson, Drew A and Manning, Christopher D},
  booktitle={CVPR},
  year={2019}
}

@inproceedings{vizwiz,
  title={Vizwiz grand challenge: Answering visual questions from blind people},
  author={Gurari, Danna and Li, Qing and Stangl, Abigale J and Guo, Anhong and Lin, Chi and Grauman, Kristen and Luo, Jiebo and Bigham, Jeffrey P},
  booktitle={CVPR},
  year={2018}
}

@inproceedings{pope,
  title={Evaluating object hallucination in large vision-language models},
  author={Li, Yifan and Du, Yifan and Zhou, Kun and Wang, Jinpeng and Zhao, Xin and Wen, Ji-Rong},
  booktitle={EMNLP},
  year={2023}
}

@article{flashattention,
  title={Flashattention: Fast and memory-efficient exact attention with io-awareness},
  author={Dao, Tri and Fu, Dan and Ermon, Stefano and Rudra, Atri and R{\'e}, Christopher},
  journal={NeurIPS},
  year={2022}
}

@article{deepseek-vl2,
  title={Deepseek-vl2: Mixture-of-experts vision-language models for advanced multimodal understanding},
  author={Wu, Zhiyu and Chen, Xiaokang and Pan, Zizheng and Liu, Xingchao and Liu, Wen and Dai, Damai and Gao, Huazuo and Ma, Yiyang and Wu, Chengyue and Wang, Bingxuan and others},
  journal={arXiv preprint arXiv:2412.10302},
  year={2024}
}

@inproceedings{vqav2,
  title={Making the v in vqa matter: Elevating the role of image understanding in visual question answering},
  author={Goyal, Yash and Khot, Tejas and Summers-Stay, Douglas and Batra, Dhruv and Parikh, Devi},
  booktitle={CVPR},
  year={2017}
}

@inproceedings{mmbench,
  title={Mmbench: Is your multi-modal model an all-around player?},
  author={Liu, Yuan and Duan, Haodong and Zhang, Yuanhan and Li, Bo and Zhang, Songyang and Zhao, Wangbo and Yuan, Yike and Wang, Jiaqi and He, Conghui and Liu, Ziwei and others},
  booktitle={ECCV},
  year={2024},
}

@article{seedbench,
  title={Seed-bench: Benchmarking multimodal llms with generative comprehension},
  author={Li, Bohao and Wang, Rui and Wang, Guangzhi and Ge, Yuying and Ge, Yixiao and Shan, Ying},
  journal={arXiv preprint arXiv:2307.16125},
  year={2023}
}

@article{mme,
  title={Mme: A comprehensive evaluation benchmark for multimodal large language models},
  author={Fu, Chaoyou and Chen, Peixian and Shen, Yunhang and Qin, Yulei and Zhang, Mengdan and Lin, Xu and Yang, Jinrui and Zheng, Xiawu and Li, Ke and Sun, Xing and others},
  journal={arXiv preprint arXiv:2306.13394},
  year={2023}
}

@article{qwen3vl,
  title={Qwen3-vl technical report},
  author={Bai, Shuai and Cai, Yuxuan and Chen, Ruizhe and Chen, Keqin and Chen, Xionghui and Cheng, Zesen and Deng, Lianghao and Ding, Wei and Gao, Chang and Ge, Chunjiang and others},
  journal={arXiv preprint arXiv:2511.21631},
  year={2025}
}

@article{llavaonevision,
  title={Llava-onevision: Easy visual task transfer},
  author={Li, Bo and Zhang, Yuanhan and Guo, Dong and Zhang, Renrui and Li, Feng and Zhang, Hao and Zhang, Kaichen and Zhang, Peiyuan and Li, Yanwei and Liu, Ziwei and others},
  journal={arXiv preprint arXiv:2408.03326},
  year={2024}
}

@inproceedings{visionzip,
  title={Visionzip: Longer is better but not necessary in vision language models},
  author={Yang, Senqiao and Chen, Yukang and Tian, Zhuotao and Wang, Chengyao and Li, Jingyao and Yu, Bei and Jia, Jiaya},
  booktitle={CVPR},
  year={2025}
}

@inproceedings{fitprune,
  title={Fit and prune: Fast and training-free visual token pruning for multi-modal large language models},
  author={Ye, Weihao and Wu, Qiong and Lin, Wenhao and Zhou, Yiyi},
  booktitle={AAAI},
  year={2025}
}

@inproceedings{fastv,
  title={An image is worth 1/2 tokens after layer 2: Plug-and-play inference acceleration for large vision-language models},
  author={Chen, Liang and Zhao, Haozhe and Liu, Tianyu and Bai, Shuai and Lin, Junyang and Zhou, Chang and Chang, Baobao},
  booktitle={ECCV},
  year={2024},
}

@inproceedings{cdpruner,
  title={Beyond attention or similarity: Maximizing conditional diversity for token pruning in mllms},
  author={Zhang, Qizhe and Liu, Mengzhen and Li, Lichen and Lu, Ming and Zhang, Yuan and Pan, Junwen and She, Qi and Zhang, Shanghang},
  booktitle={NeurIPS},
  year={2025}
}

@article{seedvl,
  title={Seed1. 5-vl technical report},
  author={Guo, Dong and Wu, Faming and Zhu, Feida and Leng, Fuxing and Shi, Guang and Chen, Haobin and Fan, Haoqi and Wang, Jian and Jiang, Jianyu and Wang, Jiawei and others},
  journal={arXiv preprint arXiv:2505.07062},
  year={2025}
}

@inproceedings{ai2d,
  title={A diagram is worth a dozen images},
  author={Kembhavi, Aniruddha and Salvato, Mike and Kolve, Eric and Seo, Minjoon and Hajishirzi, Hannaneh and Farhadi, Ali},
  booktitle={ECCV},
  year={2016},
}

@article{learnpruner,
  title={LearnPruner: Rethinking Attention-based Token Pruning in Vision Language Models},
  author={Takezoe, Rinyoichi and Li, Yaqian and Bo, Zihao and Hou, Anzhou and Guang, Mo and Long, Kaiwen},
  journal={arXiv preprint arXiv:2604.23950},
  year={2026}
}

@inproceedings{sqa,
  title={Learn to explain: Multimodal reasoning via thought chains for science question answering},
  author={Lu, Pan and Mishra, Swaroop and Xia, Tanglin and Qiu, Liang and Chang, Kai-Wei and Zhu, Song-Chun and Tafjord, Oyvind and Clark, Peter and Kalyan, Ashwin},
  booktitle={NeurIPS},
  year={2022}
}

@article{mmtok,
  title={Mmtok: Multimodal coverage maximization for efficient inference of vlms},
  author={Dong, Sixun and Hu, Juhua and Zhang, Mian and Yin, Ming and Fu, Yanjie and Qian, Qi},
  journal={arXiv preprint arXiv:2508.18264},
  year={2025}
}

@inproceedings{zoo,
  title={ZOO-Prune: Training-Free Token Pruning via Zeroth-Order Gradient Estimation in Vision-Language Models},
  author={Kim, Youngeun and Zhang, Youjia and Liu, Huiling and Jung, Aecheon and Lee, Sunwoo and Hong, Sungeun},
  booktitle={CVPR},
  year={2026}
}

@inproceedings{vispruner,
  title={Beyond text-visual attention: Exploiting visual cues for effective token pruning in vlms},
  author={Zhang, Qizhe and Cheng, Aosong and Lu, Ming and Zhang, Renrui and Zhuo, Zhiyong and Cao, Jiajun and Guo, Shaobo and She, Qi and Zhang, Shanghang},
  booktitle={ICCV},
  year={2025}
}

@inproceedings{dart,
  title={Stop Looking for “Important Tokens” in Multimodal Language Models: Duplication Matters More},
  author={Wen, Zichen and Gao, Yifeng and Wang, Shaobo and Zhang, Junyuan and Zhang, Qintong and Li, Weijia and He, Conghui and Zhang, Linfeng},
  booktitle={EMNLP},
  year={2025}
}

% \clearpage
% \appendix
% \input{appendix/appendix}


\end{document}